\documentclass{applemlr}

\usepackage{amsmath,amsfonts,bm}

\def\eqref#1{equation~\ref{#1}}

\def\1{\bm{1}}

\def\vb{{\bm{b}}}

\def\vh{{\bm{h}}}

\def\vx{{\bm{x}}}

\def\mA{{\bm{A}}}
\def\mB{{\bm{B}}}
\def\mC{{\bm{C}}}

\def\mI{{\bm{I}}}

\def\mK{{\bm{K}}}

\def\mQ{{\bm{Q}}}

\def\mV{{\bm{V}}}
\def\mW{{\bm{W}}}
\def\mX{{\bm{X}}}
\def\mY{{\bm{Y}}}

\def\mLambda{{\bm{\Lambda}}}

\DeclareMathAlphabet{\mathsfit}{\encodingdefault}{\sfdefault}{m}{sl}
\SetMathAlphabet{\mathsfit}{bold}{\encodingdefault}{\sfdefault}{bx}{n}

\newcommand{\softmax}{\mathrm{softmax}}

\usepackage[utf8]{inputenc} %
\usepackage[T1]{fontenc}    %
\usepackage{booktabs}       %
\usepackage{amsfonts}       %
\usepackage{nicefrac}       %
\usepackage{microtype}      %

\usepackage{wrapfig}
\usepackage{multirow}
\usepackage{multicol}
\usepackage{rotating}
\usepackage{fvextra}
\usepackage{float}
\newfloat{listing}{tbp}{lol}
\floatname{listing}{Listing}
\input{pygments_colorful}

\usepackage{lipsum}
\usepackage{graphicx}

\usepackage{hyphenat}
\usepackage{amssymb, amsmath, amsfonts, amsopn, array, cases, stmaryrd}
\usepackage{mathtools, xfrac, scalerel}
\usepackage{empheq}
\usepackage[makeroom]{cancel}
\usepackage[export]{adjustbox}

\usepackage{xspace}
\usepackage[textsize=tiny]{todonotes}

\newcommand*{\tran}{{\mkern-1.5mu\mathsf{T}}}      %

\DeclareMathAlphabet\mathbfcal{OMS}{cmsy}{b}{n}

\newcommand{\cmmnt}[1]{}
\newcommand{\bb}[1]{{\boldsymbol{#1}}}
\newcommand{\mc}[1]{{\mathcal{#1}}}

\usepackage{tikz,tikz-3dplot}
\usepackage{pgfplots,pgfplotstable}
\usepgfplotslibrary{patchplots,colorbrewer,fillbetween,units}
\pgfplotsset{compat=1.8}
\usepackage{arydshln,subcaption}
\usetikzlibrary{math, matrix, fadings, shapes, calc, fit, backgrounds, arrows,arrows.meta, decorations.pathreplacing, decorations.markings, positioning}
\usepackage{listofitems,siunitx,enumerate,tabu}
\pgfdeclarelayer{background}
\pgfdeclarelayer{foreground}
\pgfsetlayers{background,main,foreground}   %

\usepackage{hyperref}
\hypersetup{
  colorlinks,
  linkcolor={blue!50!black},
  citecolor={blue!50!black},
  urlcolor=black,
  breaklinks=true
}
\usepackage{cleveref}
\usepackage{enumitem}
\newlist{problems}{enumerate}{1}
\setlist[problems,1]{
    label={\textbf{Problem~\arabic*}},
    ref=\arabic*, %
    wide, labelindent=\parindent, itemsep=\parskip} 
\creflabelformat{problemsi}{#2\textup{#1}#3} %
\crefname{problemsi}{problem}{problems} %
\Crefname{problemsi}{Problem}{Problems}
\crefname{enumi}{Problem}{Problems}     %
\newlist{steps}{enumerate}{1}
\setlist[steps,1]{
    label={\textbf{Step~\arabic*}},
    ref=\arabic*, %
    wide, labelindent=\parindent, itemsep=\parskip} 
\creflabelformat{stepsi}{#2\textup{#1}#3} %
\crefname{stepsi}{step}{steps} %
\Crefname{stepsi}{step}{steps}
\crefname{enumi}{Step}{Steps}     %

\crefformat{equation}{(#2#1#3)} %

\crefname{section}{Sec.}{Sec.}
\Crefname{section}{Section}{Sections}
\crefname{appendix}{App.}{App.}
\Crefname{appendix}{Appendix}{Appendices}
\crefname{table}{Tab.}{Tab.}
\Crefname{table}{Table}{Tables}
\crefname{figure}{Fig.}{Fig.}
\Crefname{figure}{Figure}{Figures}
\crefname{algorithm}{Alg.}{Alg.}
\Crefname{algorithm}{Algorithm}{Algorithms}
\crefname{listing}{Lst.}{Lst.}
\Crefname{listing}{Listing}{Listings}

\usepackage{array}
\usepackage{bbding}
\usepackage[weather]{ifsym}
\usepackage{soul}
\sethlcolor{gray!20}
\newcommand{\band}{\rowcolor{gray!20}}
\newcolumntype{C}{>{\centering}p{5em}}

\usepackage{hyperref}
\usepackage{url}

\usepackage[most]{tcolorbox}
\usepackage{xcolor}
\definecolor{AppleWhite}{RGB}{255,255,255}
\definecolor{ApplePrimaryCoolGray}{RGB}{116,128,139}
\definecolor{AppleCoolGray1}{RGB}{199,209,214}
\definecolor{AppleCoolGray2}{RGB}{147,174,190}
\definecolor{AppleCoolGray3}{RGB}{124,147,160}
\definecolor{AppleCoolGray4}{RGB}{92,102,109}
\definecolor{AppleCoolGray5}{RGB}{78,93,100}
\definecolor{AppleCoolGray6}{RGB}{53,60,65}
\definecolor{AppleBlack}{RGB}{0,0,0}
\definecolor{AppleSecondaryChartGray}{RGB}{168,168,168}
\definecolor{AppleChartGray2}{RGB}{233,233,233}
\definecolor{AppleChartGray3}{RGB}{211,211,211}
\definecolor{AppleChartGray4}{RGB}{190,190,190}
\definecolor{AppleChartGray5}{RGB}{140,140,140}
\definecolor{AppleChartGray6}{RGB}{102,102,102}
\definecolor{AppleChartGray7}{RGB}{64,64,64}
\definecolor{ApplePrimaryChartBlue}{RGB}{84,151,193}
\definecolor{AppleBlue2}{RGB}{212,229,239}
\definecolor{AppleBlue3}{RGB}{169,202,223}
\definecolor{AppleBlue4}{RGB}{127,177,209}
\definecolor{AppleBlue5}{RGB}{71,130,166}
\definecolor{AppleBlue6}{RGB}{55,99,128}
\definecolor{AppleBlue7}{RGB}{45,72,89}
\definecolor{ApplePrimaryChartGreen}{RGB}{83,172,121}
\definecolor{AppleGreen2}{RGB}{212,234,221}
\definecolor{AppleGreen3}{RGB}{169,213,188}
\definecolor{AppleGreen4}{RGB}{126,193,155}
\definecolor{AppleGreen5}{RGB}{58,140,82}
\definecolor{AppleGreen6}{RGB}{39,102,54}
\definecolor{AppleGreen7}{RGB}{29,58,31}
\definecolor{ApplePrimaryChartYellow}{RGB}{253,195,93}
\definecolor{AppleYellow2}{RGB}{254,240,214}
\definecolor{AppleYellow3}{RGB}{254,224,174}
\definecolor{AppleYellow4}{RGB}{254,210,134}
\definecolor{AppleYellow5}{RGB}{230,168,69}
\definecolor{AppleYellow6}{RGB}{191,131,46}
\definecolor{AppleYellow7}{RGB}{153,107,54}
\definecolor{ApplePrimaryChartOrange}{RGB}{250,151,92}
\definecolor{AppleOrange2}{RGB}{254,229,214}
\definecolor{AppleOrange3}{RGB}{252,203,173}
\definecolor{AppleOrange4}{RGB}{252,178,133}
\definecolor{AppleOrange5}{RGB}{227,121,68}
\definecolor{AppleOrange6}{RGB}{191,87,46}
\definecolor{AppleOrange7}{RGB}{143,59,36}
\definecolor{ApplePrimaryChartRed}{RGB}{227,94,105}
\definecolor{AppleRed2}{RGB}{248,215,217}
\definecolor{AppleRed3}{RGB}{241,174,180}
\definecolor{AppleRed4}{RGB}{234,135,143}
\definecolor{AppleRed5}{RGB}{196,63,77}
\definecolor{AppleRed6}{RGB}{153,35,53}
\definecolor{AppleRed7}{RGB}{102,19,43}
\definecolor{ApplePrimaryChartPurple}{RGB}{161,150,204}
\definecolor{ApplePurple2}{RGB}{231,228,242}
\definecolor{ApplePurple3}{RGB}{208,202,229}
\definecolor{ApplePurple4}{RGB}{185,176,217}
\definecolor{ApplePurple5}{RGB}{128,113,171}
\definecolor{ApplePurple6}{RGB}{89,76,128}
\definecolor{ApplePurple7}{RGB}{62,46,101}
\definecolor{AppleCoolGray}{RGB}{116,128,139}
\definecolor{AppleChartGray}{RGB}{168,168,168}
\definecolor{AppleBlue}{RGB}{84,151,193}
\definecolor{AppleGreen}{RGB}{83,172,121}
\definecolor{AppleYellow}{RGB}{253,195,93}
\definecolor{AppleOrange}{RGB}{250,151,92}
\definecolor{AppleRed}{RGB}{227,94,105}
\definecolor{ApplePurple}{RGB}{161,150,204}
\definecolor{DeepBlue}{RGB}{0,114,178}
\definecolor{DeepOrange}{RGB}{230,159,0}
\definecolor{TealGreen}{RGB}{0,158,115}

\newtcolorbox{summarybox}
{
colback=AppleBlue!5,
colframe=AppleBlue5!90!black,
fonttitle=\bfseries,
title=Summary}

\definecolor{textgray}{HTML}{6E6E73}
\makeatletter
\newcommand\applefootnote[1]{%
  \begingroup
  \renewcommand\thefootnote{}%
  \renewcommand\@makefntext[1]{\noindent##1}%
  \footnote{#1}%
  \addtocounter{footnote}{-1}%
  \endgroup
}
\makeatother
\AtEndDocument{\applefootnote{\textcolor{textgray}{\sffamily Apple and the Apple logo are trademarks of Apple Inc., registered in the U.S. and other countries and regions.}}}

\title{Attention to Mamba:\\
A Recipe for Cross-Architecture Distillation}

\author[\dagger*]{Abhinav~Moudgil}
\author[\ddagger*]{Ningyuan~Huang}
\author{Eeshan~Gunesh~Dhekane}
\author{Pau~Rodr\'iguez}
\author{Luca~Zappella}
\author{Federico~Danieli}

\affiliation{Apple}
\affiliation[\dagger]{Mila -- Quebec AI Institute}
\affiliation[\ddagger]{Flatiron Institute}
\affiliation[*]{Work done during Apple internship}

\abstract{
State Space Models (SSMs) such as Mamba have become a popular alternative to Transformer models, due to their reduced memory consumption and higher throughput at generation compared to their Attention-based counterparts. On the other hand, the community has built up a considerable body of knowledge on how to train Transformers, and many pretrained Transformer models are readily available. 
To facilitate the adoption of SSMs while leveraging existing pretrained Transformers, we aim to identify an effective recipe to distill an Attention-based model into a Mamba-like architecture. In prior work on cross-architecture distillation, however, it has been shown that a na\"ive distillation procedure from Transformers to Mamba fails to preserve the original teacher performance, a limitation often overcome with hybrid solutions combining Attention and SSM blocks.
The key argument from our work is that, by equipping Mamba with a principled initialization, we can recover an overall better recipe for cross-architectural distillation. To this end, we propose a principled two-stage approach: first, we distill knowledge from a traditional Transformer into a linearized version of Attention, using an adaptation of the \emph{kernel trick}. Then, we distill the linearized version into an adapted Mamba model that does not use any Attention block.
Overall, the distilled Mamba model retains most of the original Pythia-1B Transformer performance, reaching a perplexity of 14.01 close to the teacher's 13.86. This largely carries over to downstream tasks, although some gaps appear on tasks relying more on in-context recall. To show the efficacy of our recipe, we conduct thorough ablations at 1B scale with 10B tokens varying sequence mixer architecture, scaling analysis on model sizes and total distillation tokens, and a sensitivity analysis on tokens allocation between stages.
}

\metadata[Correspondence]{\sffamily Abhinav Moudgil: \url{abhinavmoudgil95@gmail.com}}

\begin{document}

\maketitle

\section{Introduction and Motivation}
\applefootnote{This version supersedes v1, since an implementation error kept the SSM decay gates from being trained properly. We fixed it, moved the implementation from Mamba-1 to Mamba-2, and re-ran all experiments; the results reported here replace those in v1.}%
Much of the development of natural language processing over the last decade can be directly attributed to the effectiveness of the Attention mechanism \citep{bahdanau2015neural, vaswani2017attention} in generating rich, context-aware tokens representations, and unlocking parallel training. The power of Attention, however, comes at a computational cost scaling quadratically in the length of the input sequence $L$. The attempt to curb this requirement has triggered the development of a number of alternatives to Attention, which could retain linear complexity in $L$. Among these, some of the most successful are Linear Attention \citep{katharopoulos2020transformers}, RWKV \citep{peng2023rwkvreinventingrnnstransformer}, and State-Space Models (SSMs), particularly represented by Mamba \citep{gu2023mamba,dao2024transformers}.

On the one hand, the promise of faster inference times and reduced memory requirements provided by linear alternatives to Attention is undoubtedly appealing; on the other, their performance on downstream tasks still tends to fall short of that of Transformers, especially at scale. At the same time, research on Transformers is more mature, with a larger number of models available \citep{wolf-etal-2020-transformers}, and considerable computational resources already spent into pretraining said models \citep{castano2024huggingface}.
In light of this, instead of training SSMs from scratch, a promising direction is distillation \citep{hinton2015distillingknowledgeneuralnetwork}, which allows to directly leverage the knowledge embedded in readily-available pretrained Transformer models.
Na\"ive direct distillation between Transformer and Mamba architectures, however, has shown to be challenging \citep{wang2024mamballamadistillingaccelerating,bick2024transformers}, and often failing to preserve teacher performance.
In our work, we identify a critical missing piece: architectural alignment through principled initialization. Rather than forcing knowledge transfer across fundamentally different computational paradigms, we propose a two-stage bridging strategy (illustrated in \cref{fig::schematics_stages}) that exploits the mathematical connections between sequence mixers. We first distill standard Softmax Attention into Linear Attention, building up on the Hedgehog approach illustrated in \citet{zhang2024hedgehog}. This is grounded in an application of the \emph{kernel trick}, whereby the exponentials in the Attention scores computation are approximated by scalar products of specific features. The Linear Attention weights recovered after this first step are then used as an initialization for the Mamba parameters, and the whole model is further fine-tuned.
The recipe is designed to guarantee effective knowledge transfer while limiting training cost to a fraction of the one used in pretraining the teacher Transformer. As teacher models, we consider the family of pretrained Pythia Transformers \citep{biderman2023pythia}, which we distill on an SSM-adaptation of their architecture. For the distillation procedure, we use data from the OpenWebText dataset \citep{Gokaslan2019OpenWeb}. The performance of our distillation recipe is measured both in terms of sheer perplexity, and on effectiveness on downstream tasks from \texttt{lm-eval-harness} \citep{eval-harness}. Our approach retains most of the teacher model's performance: for 1B models, the student reaches a perplexity of 14.01 (from the teacher's 13.86), with good overall scores on downstream tasks. We further establish the robustness of our approach through ablations over student architecture components, a scaling analysis on model size and total distillation tokens, and a sensitivity analysis on the token budget allocation between stages.

\paragraph{Contributions} Overall, the main contributions from our work are two-fold:
\begin{itemize}
    \item We propose a novel method for cross-architecture distillation from a Transformer to a Mamba model. The method composes of two stages, whereby distillation is performed first from Attention to Linear Attention, and then onto Mamba, with the goal of favoring knowledge transfer between the two architectures.
    \item We evaluate the method effectiveness via extensive ablation, scaling and sensitivity studies. These are aimed both at refining the details of our distillation procedure, as well as verifying its robustness with respect to the available distillation budget.
\end{itemize}

\begin{figure}[t!]
    \centering
    \includegraphics[width=\linewidth,trim={0cm 0cm 0cm 0cm},clip]{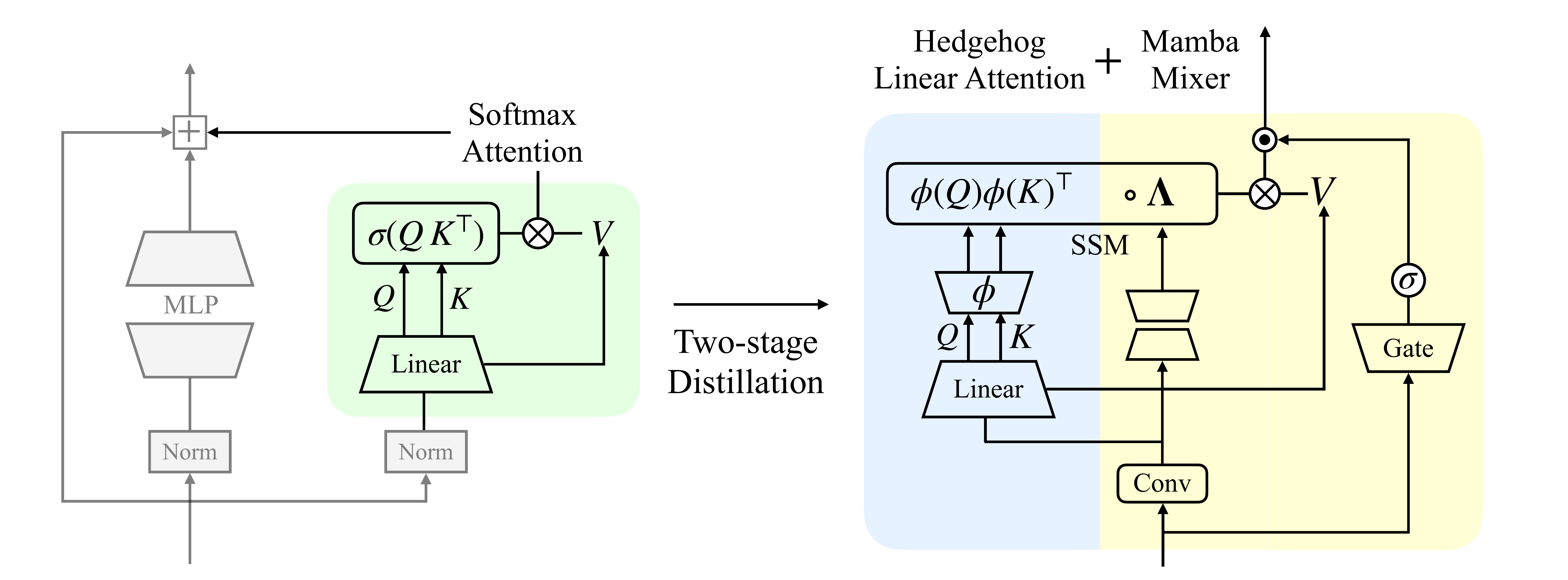}
\caption{We propose a two-stage recipe to distill quadratic Softmax Attention (green) in a Transformer layer to a subquadratic Mamba-based Mixer module. Our sequence mixer (HedgeMamba) is a hybrid of a learned linear Attention (Hedgehog~\citep{zhang2024hedgehog} in blue) and Mamba~\citep{gu2023mamba, dao2024transformers} (yellow). Note that we keep the rest of the Transformer layer as it is from the teacher model (grey); we only swap Softmax Attention with our proposed HedgeMamba mixer.}
\label{fig::schematics_attn}
\end{figure}

\subsection{Previous work}

\paragraph{Attention linearization} Attention linearization techniques aim at simplifying the operations involved in the assembly and/or application of the Attention matrix, so that their computational complexity scales \emph{linearly} (rather than \emph{quadratically}) with the sequence length. Some methods proposed in the literature achieve this by directly modifying the structure of the Attention matrix, either by sparsifying it \citep{beltagy2020longformerlongdocumenttransformer,zaheer2021bigbirdtransformerslonger} or by reducing it to low-rank \citep{wang2020linformer,xiong2021nystromformer}. Most relevant for our work is a different approach, namely kernel-based Attention linearization \citep{katharopoulos2020transformers, choromanski2022rethinkingattentionperformers, peng2023rwkvreinventingrnnstransformer,qin2022cosformerrethinkingsoftmaxattention,peng2021randomfeatureattention}. This line of research interprets the positive semi-definite Attention matrix as a kernel application, which gets decomposed as dot products of feature vectors in high-dimensional space. Effectively, this allows to reduce Attention to a Recurrent Neural Network (RNN) application. The kernel-based approach is hence of particular relevance to us as it helps bridging the gap between our two targets architectures, namely Transformers and Mamba: the latter can in fact be interpreted as a specific RNN instantiation.

\paragraph{State-Space Models} State-Space Models are a specialization of RNNs relying on recurrence formulas which are purely linear in the hidden state. Imposing linearity has the advantage of rendering the computation of the recurrence relationship parallelizable along the input sequence during training, thus overcoming one of the main limitations of classical RNNs. The line of research analyzing the properties of SSMs has been particularly active \citep{gu2020hipporecurrentmemoryoptimal,gu2022efficientlymodelinglongsequences,gu2022parameterizationinitializationdiagonalstate,cirone2025theoreticalfoundationsdeepselective}, eventually producing the Mamba architecture \citep{gu2023mamba}, which has gained particular traction as a Transformer alternative.
More recently, \citet{dao2024transformers} has drawn a direct connection between Mamba and Linear Attention (equipped with a learnable causal mask). This, together with the performance showcased by Mamba, acts as main motivation for using the Mamba architecture as a representative for Linear Attention alternatives, and for adopting it in our student model definition.

\paragraph{Cross-architecture distillation}
Knowledge Distillation \citep{hinton2015distillingknowledgeneuralnetwork} is an established method for efficiently leveraging the knowledge embedded in an already-trained \emph{teacher} model in order to accelerate the training of a \emph{student} model, with a history of successful applications \citep{Gou2021,yang2024surveyKDLLMs, mansourian2025surveyKD, busbridge2025distillationscalinglaws}. 
While the focus of most of the available literature is on distilling a teacher into a (generally smaller) student of the same model class, in our work we are interested in distilling across two \emph{different} architectures, with the purpose of reducing the computational complexity of Attention.
The literature on quadratic-to-linear Attention distillation is much less developed in this sense, but the rise of Linear alternatives to Attention has recently sparkled interest in this specific area. For example, Scavenging Hyena \citep{ralambomihanta2024scavenginghyenadistillingtransformers} distilled a Transformer model into a Hyena model \citep{poli2023hyena} (but only for small scales $<$70M); SUPRA replaced softmax Attention directly with a linear application \citep{mercat2024linearizinglargelanguagemodels}; \citet{wang2024mamballamadistillingaccelerating} proposed distillation techniques for creating efficient hybrid Transformer-Mamba models; \citet{mao2022finetuningpretrainedtransformersdecaying} simplifies distillation onto decaying fast weights, and \citep{he2025jointfinetuningconversionpretrained} studies cross-architecture alignment strategies. More recently, MOHAWK \citep{bick2024transformers,bick2025llambascalingdistilledrecurrent} has attempted Transformer-to-Mamba distillation, proposing a three-stage recipe where the output of Attention and the SSM are progressively aligned before finetuning. 
A direct quantitative comparison with MOHAWK is confounded by fundamental differences in the underlying model architectures (our work is based on Pythia, while MOHAWK utilizes Phi as backbone), and the training sets (importantly, MOHAWK uses C4 which includes Book3, a dataset known to contain copyrighted material). However, a qualitative comparison of the methodologies is instructive. Both approaches aim to harness the expressivity of Mamba-like models, but differ significantly in their design and complexity. MOHAWK employs a complex three-stage training pipeline with distinct objectives and frozen modules for each stage. By contrast, our method proposes a two-stage recipe, theoretically grounded in the functional analogies between Transformers and SSMs, offering a more direct and computationally streamlined approach.
Also relevant is LoLCATs \citep{zhang2025lolcatslowranklinearizinglarge}, which builds upon ideas from Hedgehog \citep{kasai2021finetuningpretrainedtransformersrnns, zhang2024hedgehog} (where softmax Attention is approximated via a learnable linear kernel) and aims at improving the architecture expressivity equipping it with windowed Attention and LoRA finetuning \citep{hu2021loralowrankadaptationlarge}. 
While our work also builds on Hedgehog, LoLCats is unsuitable for direct comparison due scale differences and its instruction-finetuning loss, which cannot be applied to our pretraining-style setting.

\section{Preliminaries}
\label{sec:prelims}
In this section, we provide an overview of the target architectures for our distillation procedure, namely Transformer as teacher model and Mamba as SSM student model. We also highlight the connection between linearized forms of Attention and Mamba, which we leverage in developing our distillation recipe, as detailed in \cref{sec:recipe}.

\subsection{Description of target architectures}
As representatives for the Transformer architecture, we consider models from the Pythia suite \citep{biderman2023pythia}. The suite contains publicly available models ranging in scale from 14M to 12B parameters, consistently trained following the same recipe. As target student model, we choose the Mamba architecture \citep{gu2023mamba,dao2024transformers} which arguably represents the current state-of-the-art in SSM performance. For reference, schematics of the two architectures are provided in \cref{fig::schematics}. It is worth pointing out that Mamba has been trained with the same tokenizer as Pythia, and for a similar number of steps.

At the highest level of abstraction, both the Transformer and Mamba architectures share a similar structure, which consists of interweaving two types of modules: one responsible for mixing tokens within a sequence (also called \emph{sequence mixer}), the other for mixing components of each individual token embedding (generally performed by an MLP). Arguably, the most significant difference lies in the way the sequence mixing is performed in the two architectures, as described next.

\paragraph{Attention}
In the case of Transformers, it is the Self-Attention mechanism that is responsible for mixing the tokens embeddings sequence-wise. Its action on an input sequence $\mX\in\mathbb{R}^{L \times d}$ (with $L$ being the sequence length, and $d$ the embedding dimension) is represented as
\begin{equation}
    \mY_{\text{Attn}} \coloneqq \mA_{\text{Attn}}\mV,\qquad\text{with}\qquad
    \mA_{\text{Attn}}\coloneqq\softmax\left(\frac{\mQ\mK^\tran}{\sqrt{d}}\right),
    \label{eqn:attn}
\end{equation}
where $\mQ,\mK,\mV\in\mathbb{R}^{L\times d}$ are linear transformations of $\mX$, denoting queries, keys and values.

\paragraph{SSM mixer} For Mamba, the sequence mixing is mainly performed by the SSM layer\footnote{We note that the convolution layer in Mamba applied before the SSM can also perform sequence mixing.}. This reduces to unrolling a linear recurrence relationship in the form
\begin{equation}
\begin{split}
    \vh_l &= \mLambda_l\odot \vh_{l-1} + \mB_l \otimes\mX_{l,:}\\
    \mY_{l,:} &= \mC_l^\tran \vh_l,
\end{split}\qquad\qquad
\begin{split}
    \text{for}\quad l&=1\dots L,\\
    \vh_0 &= \bb{0}\in\mathbb{R}^{N\times d},
\end{split}
\label{eqn:ssm_mixer}
\end{equation}
for parameters $\mLambda_l\in\mathbb{R}^{N\times d}$, and $\mB_l,\mC_l\in\mathbb{R}^{N}$, $N$ being the hidden state size. To highlight the similarity with \cref{eqn:attn}, the solution to the above recurrence can be expressed in matrix form as
\begin{equation}
    \mY_{\text{SSM}}\coloneqq \mA_{\text{SSM}}\mX,\qquad\text{with}\qquad
    \left[\mA_{\text{SSM}}\right]_{i,j}\coloneqq\mC_i^\tran\prod_{k=i}^{j+1}\mLambda_k\mB_j,
    \label{eqn:ssm_unrolled}
\end{equation}
which also indicates how, at training time, the SSM mixer can be applied in a parallel fashion along the sequence, similarly to Attention.
The main particularity of Mamba, which sets it apart from other SSM models, lies in the fact that the recurrence parameters $\mLambda_l,\mB_l,\mC_l$ all depend on the input $\mX_{l,:}$. Indeed, this formulation makes it akin to Linear Attention alternatives, as we outline in the following.

\paragraph{Linear Attention and SSMs}
\label{sec:prelims:linattn_ssm}
Linearized alternatives to Attention aim at turning the application of the Attention layer \cref{eqn:attn} from a quadratic- to a linear-complexity operation in $L$. One way to achieve this consists in simplifying the layer by dropping the $\softmax$ operator, to obtain
\begin{equation}
    \mY_{\text{LinAttn}}\coloneqq (\hat{\mQ}\hat{\mK}^\tran)\hat{\mV}
    =\hat{\mQ}(\hat{\mK}^\tran\hat{\mV}).
    \label{eqn:lin_attn}
\end{equation}
This simplification allows one to leverage associativity in matrix multiplication, computing first $\hat{\mK}^\tran\hat{\mV}$ and thus only instantiating much smaller matrices $\hat{\mK},\hat{\mQ},\hat{\mV}\in\mathbb{R}^{d\times L}$, rather than the full Attention matrix $\in\mathbb{R}^{L\times L}$.

By comparing \cref{eqn:lin_attn} with \cref{eqn:ssm_unrolled}, we can draw a direct connection between Linear Attention and (input-dependent) SSMs: indeed, by simplifying $\mLambda_k\equiv \mI$, one can see that the parameter $\mB,\mC,\mX$ in the SSM mixer cover a similar role as the $\hat{\mK},\hat{\mQ},\hat{\mV}$ matrices in Linear Attention. This correspondence is well-known, having been employed already in~\citet{katharopoulos2020transformers} and outlined more in detail by \citet{dao2024transformers}, and further justifies the choice of Mamba in our experiments as a representative for linearized forms of Attention. In this work, we directly leverage such correspondence to ground our distillation recipe, as described in \cref{sec:recipe:stage2}.

\section{Cross-architecture distillation}
\label{sec:recipe}
\begin{figure}[t!]
    \centering
    \includegraphics[width=\linewidth,trim={0cm 0cm 0cm 0cm},clip]{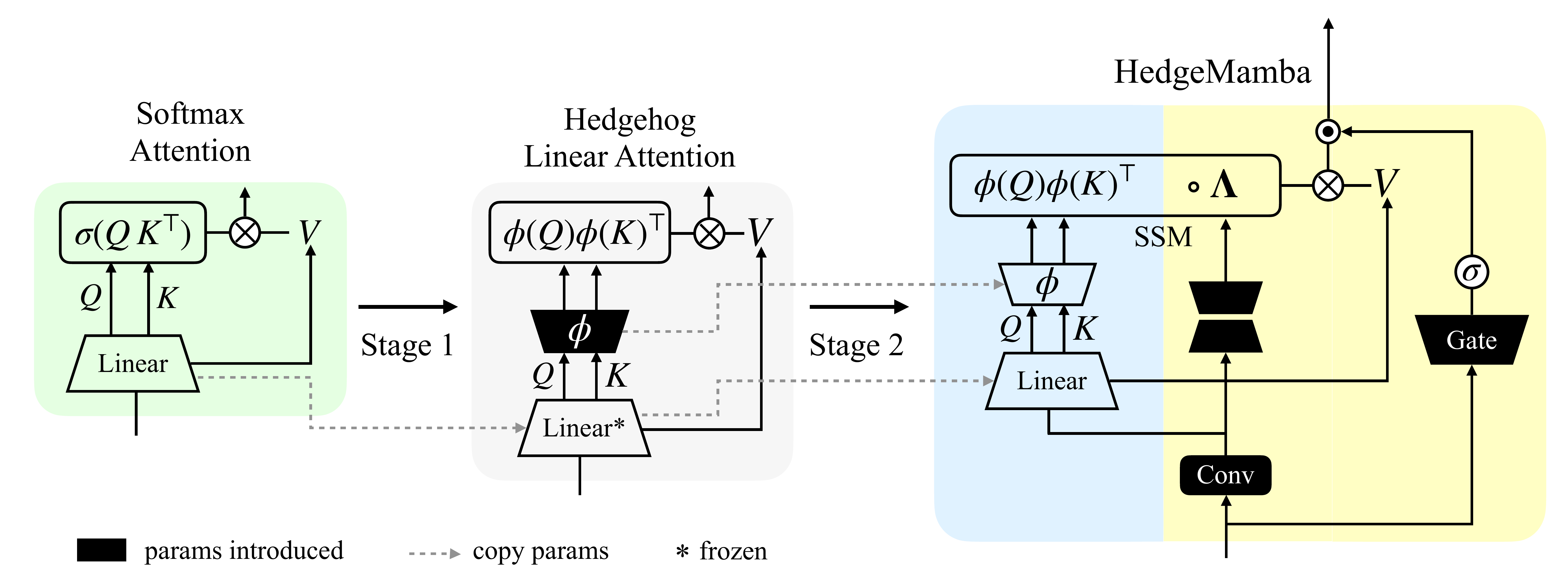}    
\caption{Schematics of the overall approach followed in our two-stage Transformer-to-Mamba distillation recipe. In stage one, we distill vanilla Attention into a linearized version, by learning a feature map $\bb{\phi}$ which approximates the action of softmax (following the Hedgehog procedure \citep{zhang2024hedgehog}). In stage two, we introduce additional components from the Mamba block, to boost the overall expressivity of the model. The resulting hybrid layer, named HedgeMamba, is then further finetuned to close the performance gap with the original teacher model.}
\label{fig::schematics_stages}
\end{figure}

In this section, we outline the distillation recipe designed and tested in this project.
The driving goal is to leverage the high-level similarity of Transformers and Mamba architectures to improve the distillation procedure.
To this end, we split our distillation recipe into two stages. In the first stage, we train a feature map to effectively distill the action of $\softmax$ Attention into Linear Attention, following the \emph{Hedgehog} procedure introduced in \cite{zhang2024hedgehog}. In the second stage, we translate the extracted Linear Attention layer into an initialization for Mamba, leveraging the correspondence outlined in \cref{sec:prelims:linattn_ssm} and \cite{dao2024transformers}, and proceed with further fine-tuning to improve overall performance. We refer to \cref{fig::schematics_stages} for an overview of each stage. %

\subsection{Stage 1: Softmax Attention to Linear Attention}
\label{sec:recipe:stage1}

The purpose of our first step is to effectively substitute $\softmax$ Attention with a linear variant that can adequately approximate its action. However, as highlighted in \citet{zhang2024hedgehog}, there is still a notable performance gap between the original $\softmax$ Attention and many existing linearizations. Motivated by this, the work in \citet{zhang2024hedgehog} focuses instead on distillation via a \emph{learnable feature map}. This is at the core of the \emph{Hedgehog} procedure introduced in \citet{zhang2024hedgehog}, which we leverage in our distillation method. We briefly define the procedure next.

\paragraph{Hedgehog} The $\softmax$ Attention scores are computed starting from various exponential terms $e^{\mQ_{l,:}\mK_{l,:}^\tran}$, but we want to remove this nonlinearity. Invoking Mercer's theorem \citep{mercer} allows us to re-write the positive definite exponential operator as a scalar product of feature vectors,
\begin{equation}
    e^{\vx^\tran\vx'} \eqcolon \kappa(\vx, \vx') = \bb{\phi}(\vx)^{\tran} \bb{\phi}(\vx'),\qquad\forall\vx,\vx'\in\mathbb{R}^d,
    \label{eqn:gauss_feature_map}
\end{equation}
for a certain feature map $\bb{\phi}(\vx):\mathbb{R}^d\to\mc{H}$.
Specifically, for the exponential kernel (also known as the Gaussian kernel), its feature space $\mathcal{H}$ is infinite-dimensional, and the feature map $\bb{\phi}(\vx)$ can be approximated via Taylor expansion of $e^z$ around $z=0$.
Linear Attention variants that aim to approximate this feature map tend to do so by keeping only the first few terms in the sum in its Taylor expansion \citep{katharopoulos2020transformers}. However, as pointed out by \citet{zhang2024hedgehog}, these variants typically do not retain some relevant features of $\softmax$ Attention, such as spikiness in the activations and dot-product monotonicity. To overcome this, \citet{zhang2024hedgehog} propose to \emph{learn} the feature map in \cref{eqn:gauss_feature_map} via a (single-layer) MLP:
\begin{equation}
    \bb{\phi}(\vx)\approx\bb{\phi}_{\texttt{MLP}}(\vx)
    \coloneq \sigma(\mW\vx + \vb),
    \label{eqn:hedgehog_MLP}
\end{equation}
with nonlinearity $\sigma$. The learnable weights $\mW\in\mathbb{R}^{d\times d},\vb\in\mathbb{R}^{d}$ are optimized by \textbf{matching the output of each teacher Attention block with that of its Hedgehog-linearized version, via cosine similarity.}
In \citet{zhang2024hedgehog}, the authors showcase how such learnable MLP feature map greatly improves the distillation performance of Linear Attention while remaining computationally efficient (we refer to their work for additional evaluation and implementation details of the Hedgehog procedure). However, while in \citet{zhang2024hedgehog} the distillation procedure stops here,
in this work we improve this approach by including the learned Hedgehog feature map into the Mamba architecture, and further refining the distillation onto this adapted architecture. More details are outlined next.

\subsection{Stage 2: Linear Attention to Mamba}
\label{sec:recipe:stage2}
With the first distillation step, we have identified a way to substitute the $\softmax$ Attention operation in \cref{eqn:attn} with a Linear Attention one. In the second step, we want to use this Linear Attention solution as an initialization to Mamba, and further fine-tune to improve the distillation performance by leveraging the additional expressivity that the Mamba module provides. In this section we show how to adapt the Mamba layer to achieve exactly this. We refer the resulting Hedgehog-Mamba layer as HedgeMamba.

\paragraph{Parameters initialization}
As mentioned in \cref{sec:prelims}, we can match the output of an SSM mixer \cref{eqn:ssm_mixer} with that of a Linear Attention layer \cref{eqn:lin_attn} by substituting $\mLambda_l\equiv\mI$ and by having the parameters $\mB,\mC,\mX$ cover a similar role as $\hat{\mK},\hat{\mQ},\hat{\mV}$. For the specific case of Hedgehog, this translates into the substitutions
\begin{equation}
\begin{array}{r@{{}\mapsto{}}lcr@{{}\mapsto{}}l}
    \mB(\mX)&\hat{\mK}(\mX)\coloneqq\bb{\phi}_{\texttt{MLP}}(\mK(\mX)),&&
    \mC(\mX)&\hat{\mQ}(\mX)\coloneqq\bb{\phi}_{\texttt{MLP}}(\mQ(\mX))\\
    \mLambda&\mI,&
    \text{and}&
    \mX&\hat{\mV}(\mX)\coloneqq\mV(\mX),
\end{array}
\label{eqn:lin_attn_2_mamba_subs}
\end{equation}
where $\mK(\mX),\mQ(\mX),\mV(\mX)$ are the key/query/value linear maps from the original $\softmax$ Attention layer \cref{eqn:attn}, while $\bb{\phi}_{\texttt{MLP}}$ is the freshly-learned Hedgehog feature map \cref{eqn:hedgehog_MLP}. Notice that the original Mamba architecture does not allow for a value transformation $\mX\mapsto\mV(\mX)$ before the application of the SSM mixer, so we modify its implementation to accommodate for this. Moreover, to ensure that the whole Mamba block output matches that of Hedgehog at initialization, we also set the parameters of the gate branch and the convolution so that they reduce to the identity operator. Additional details can be found in \cref{app::additional_details}.

\paragraph{Attention scores normalization}
With the substitution in \cref{eqn:lin_attn_2_mamba_subs}, the SSM mixer outputs
\begin{equation}
    \mY_{\phi} \coloneqq 
    \left(\bb{\phi}_{\texttt{MLP}}(\mQ)\bb{\phi}_{\texttt{MLP}}(\mK)^\tran\right)\mV.
    \label{eqn:softmax-linear-learnable}
\end{equation}
However, the Attention scores in this formula come in an un-normalized fashion. For the Attention scores formulation to more closely follow the target one in \cref{eqn:attn}, we further include a normalization factor in their definition, as in the original Linear Attention formulation~\citep{katharopoulos2020transformers}:
\begin{equation}
    \mY_{\phi} \mapsto \mY_{\phi}/\bar{\mY}_{\phi},\qquad\text{with}\qquad
    \bar{\mY}_{\phi} \coloneqq 
    \left(\bb{\phi}_{\texttt{MLP}}(\mQ)\bb{\phi}_{\texttt{MLP}}(\mK)^\tran\right)\boldsymbol{1}.
    \label{eqn:state_normalisation}
\end{equation}
Notice that both $\mY_{\phi}$ and $\bar{\mY}_{\phi}$ can be computed with a single pass through the SSM mixer, provided we expand $\mV$ with an all-one vector\footnote{For Mamba-1, where $\mLambda$ varies per-channel, normalization would require duplicating also the state matrix $\mLambda\mapsto\operatorname{concat}[\mLambda;\mLambda]$, while form Mamba-2 this is not necessary, as the state matrix is a scalar per-head.}, that is
\begin{equation}
    \mV\mapsto\operatorname{concat}[\mV;\bb{1}].
    \label{eqn:state_expansion}
\end{equation}
In practice, we did not observe instabilities during training, but the additional normalization step provided some marginal improvements during preliminary experiments.

\paragraph{Fine-tuning} With Mamba initialized as in \cref{eqn:lin_attn_2_mamba_subs}, and modified to accommodate for normalization as per \cref{eqn:state_normalisation} and \cref{eqn:state_expansion}, we are ready to resume training and enter the second stage of our distillation procedure. This amounts to \textbf{fine-tuning the whole architecture} (except the embedding layers) \textbf{via Cross-Entropy loss with respect to the ground-truth}. Particularly, we also unlock the additional convolution and gate branches available in the original Mamba block, completing our definition of the HedgeMamba layer: see also \cref{fig::schematics_stages} for an outline of its components fine-tuned in this final stage.

The key argument from our work is that by equipping Mamba with a Hedgehog initialization we can recover an overall better recipe for cross-architectural distillation. As we illustrated in this section, our two-stage approach is theoretically grounded in Mercer's theorem, together with the superior expressive power of Mamba over vanilla Linear Attention. In the following section, we proceed to justify our approach also empirically, by benchmarking models trained following our recipe.

\section{Results}
\label{sec::results}
In this section, we present an extensive evaluation of the distillation procedure outlined in \cref{sec:recipe}. Specifically, we conduct ablation, scaling, and sensitivity studies with Pythia-1B teacher model and 10B distillation tokens across several key axes: (i) mixer architecture, by systematically expanding vanilla Hedgehog Linear Attention with components from Mamba~(\cref{tab::varying_mamba_comp}); (ii) sensitivity analysis on token budget allocation in different stages~(\cref{tab::token_budget_alloc}); and (iii) scaling with respect to number of distillation tokens~(\cref{tab::scaling_distil_tokens}). Our default settings are \hl{highlighted} in their respective tables. 
In~\cref{app::additional_results} we further expand on the results in this section, including results on applying our distillation procedure to different model sizes (160M, 410M, and 1B), and reporting standard error for results in~\cref{tab::varying_mamba_comp,tab::token_budget_alloc,tab::scaling_distil_tokens}.

\paragraph{Experimental setup} For all experiments we use standard Pythia-1B~\citep{biderman2023pythia} as teacher model. This model has been widely adopted by the open source community, and the suite provides both the model weights (for different scales), and the detailed full training procedure. We distill our models using the  OpenWebText~\citep{Gokaslan2019OpenWeb} dataset. This is an open-source reproduction of the dataset used to train GPT2~\citep{radford2019language}, and it is commonly used in language modeling research~\citep{biderman2023pythia, sanh2019distilbert, dao2022flashattention, shoeybi2019megatron, zhuang-etal-2021-robustly}. We employ the same GPT-NeoX tokenizer used for the original Pythia and Mamba models, making our results directly comparable. This amounts to a total of about 9B tokens available for training. We keep a $0.05\%$ split on the dataset for validation, which corresponds to 4M tokens, as in prior works~\citep{dao2022flashattention}. Unless otherwise reported, we use a total of 10B tokens  (roughly corresponding to 1.1 epochs of OpenWebText) which, to the best of our knowledge, establishes our work as the currently largest sensitivity study with respect to token budget in distillation. We evaluate the final distilled student models both in terms of upstream perplexity as well as performance on selected downstream tasks. For the latter, we rely on the \texttt{lm-eval-harness}~\citep{eval-harness} test suite, and consider language understanding and common sense reasoning used in prior work~\citep{biderman2023pythia, gu2023mamba, dao2024transformers}. More specifically, we report accuracy scores for ARC-Easy~\citep{clark2018think}, Social IQA~\cite{sap-etal-2019-social}, PiQA~\cite{bisk2020piqa}, Lambada~\citep{lambada}, BoolQ~\citep{boolq}, RACE~\citep{race}, LogiQA~\citep{logiqa}, and WinoGrande~\citep{winogrande}, and accuracy normalized by sequence length for ARC-Challenge~\citep{clark2018think} and HellaSwag~\citep{hellaswag}, as in~\citep{bick2024transformers, gu2023mamba, dao2024transformers, sanh2019distilbert}. We refer the reader to~\citep{eval-harness} for more details regarding evaluation.

\paragraph{Training}
In the first stage of recipe (\cref{sec:recipe:stage1}), we replace the Attention block in the teacher model with the Hedgehog linearization, with the goal of learning the feature map \cref{eqn:hedgehog_MLP}. Except for the parameters defining this feature map (which are learned from scratch in stage 1), all the other parameters are copied directly from the teacher model and kept frozen. These include MLPs, layer norms, and input-output embedding matrices. We match the output of each Transformer layer (consisting of MLP, sequence mixer, and residual stream) in the student model with those from the teacher via cosine embedding matching loss. We use 1B tokens for stage 1 of our distillation recipe with batch size 48 and sequence length 1024, which corresponds to 20K training steps. In the second stage (\cref{sec:recipe:stage2}), we introduce additional Mamba parameters, initialized to the identity operator (see \cref{app::additional_details}). We keep the input-output embedding layers frozen, and finetune the rest of the model with standard cross-entropy loss. Second-stage training continues for another 9B tokens, corresponding to additional 180K training steps.

\paragraph{Implementation remarks} We use the teacher models implementations and pretrained weights directly from the HuggingFace Transformers library~\citep{wolf-etal-2020-transformers}. Student models are implemented by swapping the softmax Attention modules from the teacher with Mamba Mixer modules from~\cite{dao2024transformers}, equipped with the Hedgehog feature maps from~\cite{zhang2024hedgehog}. For the HedgeMamba layer in \cref{fig::schematics_stages} (see the corresponding code in \cref{app::pseudocode}), we build on Mamba-2~\citep{dao2024transformers}, despite it employing in its recurrence a less general formulation than the original Mamba~\citep{gu2023mamba}. The main difference between the two SSMs lies in the parameterization of their respective state transition matrices: for Mamba-2, $\mLambda$ is scalar-times-identity, while for Mamba-1 this is diagonal, hence strictly more expressive. On the other hand, however, the selective scan kernel at the core of Mamba-1 application is much less developed than its twin from Mamba-2: it is tuned for small state sizes (16 by default), it does not exploit tensor cores, and it does not support state sizes above 256~\citep{mambahardcapissue}, whereas HedgeMamba operates at 2048. Mamba-1 can still be used, serializing its application, but initial experiments showed no gain from the additional expressivity, the final perplexity of the distilled Mamba-1 model falling within noise with respect to the Mamba-2 one. For this reason, we focused on Mamba-2 in the rest of this paper. With our default settings, distilling Pythia-1B with 10B tokens on a single 8$\times$A100 node takes 2.8h for stage 1 (20K steps) and 29.8h for stage 2 (180K steps, at roughly 0.597s/step). For reference, a stage 2 step of the Hedgehog baseline takes 0.657s, on par with our approach. Further implementation details are in \cref{app::implementation_details}. 

\paragraph{Baseline comparison}
    
\begin{table*}[b!]
\caption{Comparison with teacher and prior work.}
\label{tab::eval_baselines}
\footnotesize
\centering
\resizebox{\textwidth}{!}{\begin{tabular}{l<{\hspace{-5pt}} c<{\hspace{-5pt}} *{10}{c<{\hspace{-7.5pt}}}}
    \toprule
    Model (1B) & $\downarrow$~PPL  & $\uparrow$~Arc-C & Arc-E  & SIQA & PiQA
    &Lambada &BoolQ &RACE &LogiQA &WinoG & HSwag \\
    \midrule
    Pythia (Teacher)         & 13.86 & 27.05 & 56.99 & 39.87 & 70.73 & 42.07 & 60.82 & 32.92 & 22.12 & 53.43 & 47.16\\
    Hedgehog (Baseline)      & 14.89 & 26.45 & 52.74 & 38.38 & 68.01 & 30.60 & 54.80 & 30.43 & 21.66 & 50.91 & 40.79\\
    \band HedgeMamba (Ours)  & \textbf{14.01} & 27.22 & 53.24 & 40.23 & 68.72 & 32.74 & 55.38 & 30.62 & 23.20 & 51.85 & 42.38\\
    \bottomrule
\end{tabular}}
\end{table*}
Our main objective is to showcase the ability of our two-stage recipe to retain the performance of the original teacher model. We point out that this is not guaranteed, given the substantial differences with the teacher's architecture, and the fact that we consider a pure Linear Attention variant---and not hybrids like in some prior works~\citep{wang2024mamballamadistillingaccelerating, bick2024transformers}. Moreover, the original Pythia-1B teacher model was trained with 300B tokens~\citep{biderman2023pythia}, but we distill it with \textbf{only 10B, corresponding to $\sim$3.0\% of the token budget used to train the teacher}.
The distillation results are reported in \cref{tab::eval_baselines}. We compare our recipe against the Hedgehog baseline, distilled using the same cosine matching objective in stage 1 and cross-entropy in stage 2, although in this case the split in distillation tokens is 50/50 among the two stages, as per their original work~\citep{zhang2024hedgehog, zhang2025lolcatslowranklinearizinglarge}.
Overall, our approach manages to retain the teacher performance for the large part, scoring a perplexity of 14.01 versus the original 13.86.
The largest gap appears on Lambada (32.74 versus 42.07). We attribute it to Attention linearization itself: in about 80\% of Lambada passages the target word already appears in the context \citep{lambada}, so the task rewards the ability to recall tokens seen earlier in the input, a known weak spot of Linear Attention compared to softmax Attention \citep{arora2024simple}. Even in this case, though, we recover part of the gap with respect to the Hedgehog baseline (30.60).
We also point out that our approach outperforms the scaled-up Hedgehog baseline with 10B tokens in both upstream and downstream performance, highlighting the efficacy of our approach. As additional baseline (not reported here), we tested against na\"ive direct distillation onto a Mamba architecture, but consistently resulted in unsatisfactory results (PPL>100), confirming the findings from \citet{bick2024transformers}.

\paragraph{Ablating Mamba components}
\label{sec:ablating}
\begin{table*}[t!]
\caption{Mixer architecture ablations: perplexity on validation set and downstream task performance.}
\label{tab::varying_mamba_comp}
\scriptsize
\centering
\resizebox{\textwidth}{!}{
\begin{tabular}{l *{2}{c<{\hspace{-5pt}}} *{10}{c<{\hspace{-7pt}}}}
    \toprule
    Model (1B) & \#params & $\downarrow$~PPL  & $\uparrow$~Arc-C &Arc-E  &SIQA &PiQA
    &Lambada &BoolQ &RACE &LogiQA &WinoG & HSwag\\
    \midrule
    Hedgehog                             & 1,014M & 14.89 & 26.45 & 52.74 & 38.38 & 68.01 & 30.60 & 54.80 & 30.43 & 21.66 & 50.91 & 40.79\\
    \midrule
    +SSM                                 & 1,020M & 14.75 & 26.62 & 52.27 & 38.74 & 69.21 & 32.95 & 53.39 & 29.67 & 22.43 & 50.83 & 41.07\\
    \hspace{3pt}+Conv                    & 1,020M & 14.75 & 26.62 & 53.16 & 38.84 & 68.23 & 32.10 & 55.11 & 30.33 & 23.20 & 50.83 & 41.19\\
    \band \hspace{6pt}+Gate(HedgeMamba)  & 1,087M & \textbf{14.47} & 26.96 & 53.83 & 38.59 & 69.21 & 32.68 & 55.47 & 30.53 & 22.73 & 51.78 & 41.70\\
    \bottomrule
\end{tabular}}
\end{table*}
In the ablation study in \cref{tab::varying_mamba_comp}, we investigate the role of the additional Mamba components included in stage 2 in improving the final performance of the student model. Specifically, we consider simple Hedgehog as a baseline~\citep{zhang2024hedgehog}, and systematically add the following components from Mamba: (+SSM) the SSM mixer parameters, particularly the learnable causal mask $\bb{\Lambda}$ and input and output matrices $\mC$ and $\mB$ from \cref{eqn:ssm_mixer}; (+Conv) the short-convolution layer at the input; (+Gate) the gate branch with \texttt{SiLU} non-linearity.
These added components are initialized to behave like the identity, not to affect the Hedgehog feature map learned in stage 1 (see \cref{app::additional_details}). The other SSM mixer parameters $\mC$ and $\mB$ are instead directly copied from their equivalent in the Hedgehog module, as described in \cref{eqn:lin_attn_2_mamba_subs}.
To analyze the impact of the new introduced mixer components in a targeted manner, all the other ablation parameters are kept fixed: particularly, we use 10B distillation tokens, split evenly (50/50) among stages 1 and 2. The corresponding results are reported in \cref{tab::varying_mamba_comp}. There, we can see how each additional Mamba component contribute to improving the performance of vanilla Hedgehog. The largest improvements in perplexity and average downstream performance are yielded by the learnable SSM causal mask and the gate branch: the first finding further motivates our distillation approach, while the latter is consistent with recent works~\citep{qiu2025gatedattentionlargelanguage, pmlr-v162-hua22a, bondarenko2023quantizable}, that suggest adding a gate branch to Attention modules to improve their performance. The convolution contribution instead improves average downstream performance only marginally.

\begin{table*}[b!]
  \caption{Sensitivity analysis on the token budget allocation. Total 10B distillation tokens. Distillation on full HedgeMamba, with convolution layer and gate branch. Notice that a 100/0 split indicates that no fine-tuning is performed, and all tokens are used for learning the Hedgehog map in stage 1 (S1); conversely, 0/100 indicates only fine-tuning (S2) on the HedgeMamba architecture.}
  \label{tab::token_budget_alloc}
  \setlength{\tabcolsep}{0.4em}
  \tiny
  \resizebox{\textwidth}{!}{
\begin{tabular}{l c c<{\hspace{-5pt}} *{10}{c<{\hspace{-5pt}}}}
    \toprule
    & \textbf{Tokens split} & & \multicolumn{10}{c}{\bf $\uparrow$~Downstream evals} \\
    \cmidrule(lr){2-2}
    \cmidrule(lr){4-13}
    & S1 / S2 (\%) & $\downarrow$~PPL & Arc-C & Arc-E  & SIQA & PiQA  & Lambada & BoolQ & RACE & LogiQA & WinoG & HSwag\\
    \midrule
    Hedgehog (no FT)  & 100 / 0 & 25.71 & 25.85 & 48.70 & 36.34 & 66.49 & 12.12 & 61.47 & 27.27 & 20.58 & 50.83 & 26.14\\
                      & 90 / 10 & 15.99 & 25.34 & 52.53 & 38.59 & 68.66 & 29.69 & 53.52 & 30.14 & 21.97 & 51.62 & 39.93\\
                      & 75 / 25 & 15.08 & 26.62 & 52.82 & 39.00 & 68.88 & 32.06 & 61.16 & 30.81 & 21.97 & 50.83 & 40.91\\
                      & 50 / 50 & 14.47 & 26.96 & 53.83 & 38.59 & 69.21 & 32.68 & 55.47 & 30.53 & 22.73 & 51.78 & 41.70\\
                      & 25 / 75 & 14.16 & 27.05 & 52.53 & 38.59 & 68.61 & 33.13 & 57.98 & 30.81 & 23.04 & 50.51 & 42.03\\
    \band Default     & 10 / 90 & \textbf{14.01} & 27.22 & 53.24 & 40.23 & 68.72 & 32.74 & 55.38 & 30.62 & 23.20 & 51.85 & 42.38\\
    Finetune only     & 0 / 100 & 15.74 & 25.94 & 51.81 & 38.28 & 67.95 & 25.54 & 56.51 & 27.85 & 23.50 & 51.93 & 39.05\\
    \bottomrule
    \end{tabular}}
\end{table*}
\paragraph{Token budget allocation between stages}
One relevant design choice for our recipe consists in determining how to best allocate our distillation tokens budget among the two stages. In \cref{tab::token_budget_alloc} we verify this empirically, and report student performance evaluations when varying how the distillation tokens are split between stages 1 and 2. Notice that in the original Hedgehog paper~\cite{zhang2024hedgehog}, the authors settle for a 50/50 split; our results instead show that it is progressively more beneficial to invest into stage 2, up to $90\%$ of the total token budget. 
Still, both stages are needed to guarantee performance, as testified by the poor results attained by the extreme 100/0 and 0/100 splits.

\paragraph{Scaling number of distillation tokens}
For reference, in \cref{tab::scaling_distil_tokens} we report how the final performance of our student model scales with respect to the number of distillation tokens available. We distill Pythia-1B onto full HedgeMamba, with convolution layer and gate branch. The tokens budget split between the two stages is fixed at the optimal 10/90, and we only vary total budget. Overall, student perplexity improves as the token budget increases, and at 10B it has not yet reached saturation.

\begin{table*}[t!]
\caption{Scaling study on the distillation token budget.}
\label{tab::scaling_distil_tokens}
\small
\centering
\resizebox{\textwidth}{!}{
\begin{tabular}{l c *{12}{c<{\hspace{-7pt}}}}
    \toprule
    Token budget & $\downarrow$~PPL & $\uparrow$~Arc-C & Arc-E  & SIQA & PiQA  & Lambada & BoolQ & RACE & LogiQA & WinoG & HSwag\\
    \midrule
    1B         & 16.34 & 25.77 & 50.84 & 38.43 & 68.12 & 27.19 & 55.90 & 29.09 & 22.73 & 51.30 & 39.81\\
    2B         & 15.47 & 26.02 & 51.43 & 38.54 & 69.21 & 29.73 & 58.53 & 29.67 & 21.81 & 50.99 & 40.54\\
    3B         & 15.05 & 25.68 & 52.78 & 38.64 & 69.53 & 30.91 & 58.81 & 29.19 & 20.74 & 50.83 & 41.06\\
    \band 10B  & \textbf{14.01} & 27.22 & 53.24 & 40.23 & 68.72 & 32.74 & 55.38 & 30.62 & 23.20 & 51.85 & 42.38\\
    \bottomrule
\end{tabular}}
\end{table*}

\paragraph{Inference efficiency}
\begin{table*}[b!]
\caption{Inference efficiency of HedgeMamba and of its Pythia-1B teacher on a single A100-80GB GPU, for varying sequence length $L$ and batch size $B$. Left: cache memory (model weights and activations excluded). Middle: decoding throughput over 128 generated tokens. Right: best throughput over the tested batch sizes (corresponding $B$ in brackets), and resulting speedup of HedgeMamba over Pythia. ($\times$ run out of memory.)}
\label{fig::efficiency}
\centering
\footnotesize
\setlength{\tabcolsep}{3pt}
\setlength{\aboverulesep}{0pt}\setlength{\belowrulesep}{0pt}\renewcommand{\arraystretch}{1.3}
\begin{adjustbox}{max width=\textwidth}
\begin{tabular}[t]{c l *{5}{r} || *{5}{r}}
    \toprule
    && \multicolumn{5}{c}{Cache memory [MiB]} & \multicolumn{5}{c}{Throughput [Tkn/s]}\\
    \cmidrule(lr){3-7}\cmidrule(lr){8-12}
    & $L{\downarrow}$ & $B{\rightarrow}$1 & 4 & 8 & 16 & 32 & 1 & 4 & 8 & 16 & 32\\
    \midrule
    \multirow{5}{*}{\rotatebox[origin=c]{90}{\scriptsize HedgeMamba}}
    & 1k  & 32 & 130 & 259 & 518 & 1036 & 26 & 104 & 207 & 432 & 860\\
    & 2k  & 32 & 130 & 259 & 518 & 1036 & 24 & 104 & 206 & 430 & 856\\
    & 4k  & 32 & 130 & 259 & 518 & 1036 & 24 & 105 & 215 & 427 & 849\\
    & 8k  & 32 & 130 & 259 & 518 & 1036 & 24 & 105 & 204 & 428 & 848\\
    & 16k & 32 & 130 & 259 & 518 & 1036 & 25 & 105 & 207 & 421 & 850\\
    \midrule
    \multirow{5}{*}{\rotatebox[origin=c]{90}{\scriptsize Pythia}}
    & 1k  & 128 & 512 & 1024 & 2048 & 4096 & 66 & 252 & 506 & 1034 & 1313\\
    & 2k  & 256 & 1024 & 2048 & 4096 & 8192 & 56 & 221 & 486 & 634 & 746\\
    & 4k  & 512 & 2048 & 4096 & 8192 & $\times$ & 55 & 243 & 361 & 353 & $\times$\\
    & 8k  & 1024 & 4096 & 8192 & $\times$ & $\times$ & 53 & 186 & 190 & $\times$ & $\times$\\
    & 16k & 2048 & 8192 & $\times$ & $\times$ & $\times$ & 54 & 100 & $\times$ & $\times$ & $\times$\\
    \bottomrule
\end{tabular}%
\hspace{2em}%
\begin{tabular}[t]{l ccc}
    \toprule
    & \multicolumn{2}{c}{Best throughput [Tkn/s]} & \\
    \cmidrule(lr){2-3}
    $L{\downarrow}$ & HedgeMamba & Pythia & Speedup\\
    \midrule
    1k  & 860 (32) & 1313 (32) & 0.66$\times$\\
    2k  & 856 (32) & 746 (32) & 1.15$\times$\\
    4k  & 849 (32) & 361 (8) & 2.35$\times$\\
    8k  & 848 (32) & 190 (8) & 4.47$\times$\\
    16k & 850 (32) & 100 (4) & 8.51$\times$\\
    \bottomrule
\end{tabular}
\end{adjustbox}
\end{table*}
The main motivation behind our goal of replacing Attention is to attain faster, cheaper inference: here we measure this gain, comparing the inference cost of the distilled HedgeMamba student with that of its Pythia-1B teacher. To this end, we prefill $B$ random prompts of length $L$, and greedily decode 128 tokens, one at a time, carrying the cache forward; throughput is measured over the decoding loop only, excluding prefill and warm-up iterations. All measurements are taken on a single A100-80GB GPU: for Pythia, we using its HuggingFace implementation with FlashAttention-2~\citep{dao2023flashattention2}; for HedgeMamba, we rely on Mamba's fast decoding path~\citep{dao2024transformers}. Results are reported in \cref{fig::efficiency}. As expected, the cache memory of HedgeMamba (consisting of its recurrent and convolution states) is constant in $L$, whereas the KV cache of Pythia grows linearly in $L$, eventually running out of memory (left in \cref{fig::efficiency}). Similarly, the throughput of HedgeMamba is unaffected by $L$, while that of Pythia decreases (middle in \cref{fig::efficiency}). At small batch sizes Pythia still remains faster, but this comparison disregards available memory: in a realistic deployment, the batch size would be adjusted to maximize throughput within the available memory. Under this criterion (right in \cref{fig::efficiency}), HedgeMamba is slower than Pythia only at $L=1$k (0.66$\times$), and is already 1.15$\times$ faster at $L=2$k, up to 8.51$\times$ at $L=16$k. In short, at a $16$k context length, the same GPU can serve 8$\times$ more tokens per second, with a gap that keeps widening as the context grows: this is precisely the payoff that motivates our work.

\section{Conclusion}
\label{sec::conclusion}
In this paper, we propose a novel recipe to distill a Transformer model into an SSM-based architecture. The goal is to allow the user to reduce inference time (from quadratic in sequence length, as in classical $\softmax$ Attention, to linear), without having to train a new architecture from scratch, and instead aptly leveraging already-available pretrained models. The design of our architecture relies on a principled approach, first approximating $\softmax$ Attention with a Linear Attention variant, and then use it to initialize a Mamba block, to boost its expressivity. Mirroring these steps, the distillation procedure is also composed of two stages: the first with the goal of aligning Attention weights, and the second allowing for further fine-tuning of the whole architecture. In particular, the inclusion of this first stage has proven to boost outputs alignment between student and teacher, over na\"ive direct distillation. The effectiveness of the procedure is evaluated both in terms of perplexity and performance on downstream tasks, showing that the student retains most of the teacher performance, although downstream tasks relying more on in-context recall tend to show larger gaps.

\paragraph{Limitations and future work}
To maintain a clear focus, we targeted our analysis on a specific Transformer architecture (namely, Pythia). In principle, our recipe is flexible enough to be extended to other Attention-based models, but we have not investigated its effectiveness on other variants. Also, we do not isolate the effect of distillation dataset quality on the final student performance, and limit our experiments to OpenWebText only. We leave both of these as relevant directions for future research.
Finally, in the scope of this work, we investigated \emph{one} way of boosting the student model expressivity, that is by incorporating components from the Mamba architecture: the space of possible extensions, however, remains open to additional exploration which might further increase final performance. This notwithstanding, we believe that our work represents a meaningful step in this exploration, covering a previously unexplored approach to bridging the gap between Attention and Mamba.

\newpage

\bibliography{arxiv2026/bibliography}
\bibliographystyle{iclr2026/iclr2026_conference}

\newpage

\appendix
\crefalias{section}{appendix}
\crefalias{subsection}{appendix}

\clearpage
\section{Additional results and implementation details}

\subsection{Expanded Results}
\label{app::additional_results}

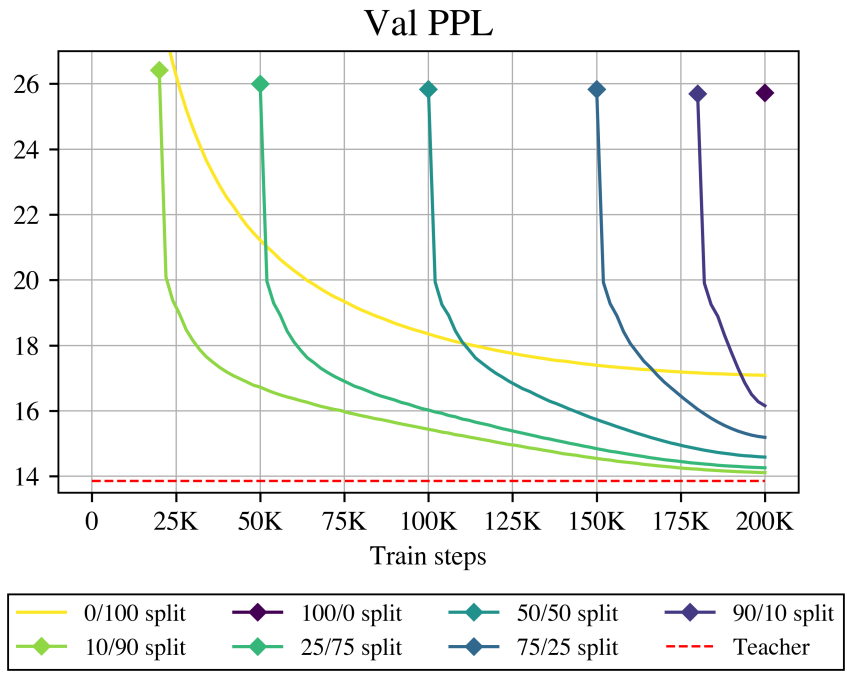
\begin{figure}[b!]
\centering
\definecolor{cA}{RGB}{223,227,24}    %
\definecolor{cB}{RGB}{117,208,84}    %
\definecolor{cC}{RGB}{42,176,127}    %
\definecolor{cD}{RGB}{35,138,141}    %
\definecolor{cE}{RGB}{50,100,142}    %
\definecolor{cF}{RGB}{69,55,129}     %
\definecolor{cG}{RGB}{68,1,84}       %

\newcommand{\datafile}{figs/fig3_token_split.dat}

\newcommand{\splitcurve}[3]{%
  \addplot[thick, #1, mark=none]
      table[x=step, y expr={\thisrow{split}==#2 ? \thisrow{ppl} : nan}] {\datafile};%
  \addplot[#1, only marks, mark=diamond*, mark size=3.4pt, forget plot]
      table[x=step,
            y expr={(\thisrow{split}==#2 && \thisrow{start}>0) ? \thisrow{ppl} : nan}]
      {\datafile};%
  \addlegendentry{#3}%
}

\newcommand{\splitline}[4]{%
  \addplot[thick, #1, mark=none, restrict y to domain=13.2:27.5]
      table[x=step,
            y expr={(\thisrow{split}==#2 && \thisrow{lr}==#3) ? \thisrow{ppl} : nan}]
      {\datafile};%
  \addlegendentry{#4}%
}

\newcommand{\splitmarker}[3]{%
  \addplot[#1, only marks, mark=diamond*, mark size=3.4pt]
      table[x=step, y expr={\thisrow{split}==#2 ? \thisrow{ppl} : nan}] {\datafile};%
  \addlegendentry{#3}%
}

\begin{tikzpicture}
\begin{axis}[
    width=12cm, height=8cm,
    xlabel={Total training steps (stage 1 + stage 2)},
    ylabel={Validation perplexity},
    xmin=0, xmax=208000,
    ymin=13.2, ymax=27.5,
    xtick={0,50000,100000,150000,200000},
    xticklabels={0,50k,100k,150k,200k},
    scaled x ticks=false,          %
    ytick={14,16,18,20,22,24,26},
    grid=major,
    grid style={gray!22, line width=0.3pt},
    tick align=outside,
    tick label style={font=\small},
    label style={font=\small},
    legend style={
        at={(0.5,-0.24)}, anchor=north,
        legend columns=4,
        font=\small,
        draw=gray!50,
        /tikz/every even column/.append style={column sep=8pt},
    },
    legend cell align=left,
    reverse legend,
]
\addplot[black!55, dashed, line width=0.9pt, mark=none, domain=0:208000, samples=2]
    {13.867};
\addlegendentry{Teacher}
\splitline{cA, line width=1.3pt}{0}{1}{0/100}
\splitcurve{cB, line width=1.1pt}{10}{10/90}
\splitcurve{cC}{25}{25/75}
\splitcurve{cD}{50}{50/50}
\splitcurve{cE}{75}{75/25}
\splitcurve{cF}{90}{90/10}
\splitmarker{cG}{100}{100/0}
\node[anchor=south west, font=\scriptsize, black!55]
    at (axis cs:3000,13.9) {Pythia-1B, 13.87};
\end{axis}
\end{tikzpicture}
\caption{Visualizing evolution of validation perplexity during stage 2 training, for different stage1/stage2 token allocation splits. $\blacklozenge$ indicates the PPL value at completion of stage 1. All sensitivity studies were run for 200K training steps, corresponding to roughly 10B tokens.}
\label{fig::plot_val_ppl}
\end{figure}

\begin{summarybox}
In this section we expand the results in \cref{sec::results} and find that:
\begin{itemize}
    \item Increasing the scale of the architectures (from 160M, to 410M and 1B) reduces PPL and increases performance, as shown in~\cref{tab::scaling_baselines_exp}.
    \item Even though stage 2 provides the larger benefits in decreasing overall PPL, the role of stage 1 is key in improving performance, as shown in \cref{fig::plot_val_ppl}.
\end{itemize}

\end{summarybox}

In this section we expand upon the results provided in \cref{sec::results}.
Particularly, \cref{tab::eval_baselines_exp} collects the results in \cref{tab::varying_mamba_comp,tab::token_budget_alloc,tab::scaling_distil_tokens}, and reports also their associated standard error over 100,000 bootstrap repetitions (as per default \texttt{lm-eval} settings\footnote{LM harness evaluation estimates the Standard Error of the Mean (SEM) using bootstrap resampling: it repeatedly samples a set of multiple choice questions (for a specified number of bootstrap iterations, 100K in our case) and then calculates the SEM of the metric scores obtained from these samples. Relevant code for this error computation can be found in \href{https://github.com/EleutherAI/lm-evaluation-harness/blob/c5acce0c7485a3bc8cbb7d80bf6a577a6922bafa/lm_eval/api/metrics.py\#L445}{\texttt{https://github.com/EleutherAI/lm-evaluation-harness}}.}). Overall, standard deviation remains low across all tasks, indicating their robustness.

\begin{table*}[b!]
\caption{Results from experiments in \cref{sec::results}, including standard error for LM evaluations.}
\label{tab::eval_baselines_exp}
\small
\centering
\resizebox{\textwidth}{!}{
\begin{tabular}{l *{10}{c<{\hspace{-7pt}}}}
    \toprule
    Model (1B) & $\uparrow$~Arc-C &Arc-E  &SIQA &PiQA
    &Lambada &BoolQ &RACE &LogiQA &WinoG & HSwag\\
    \midrule
    Pythia(Teacher)                     & 27.05\tiny{1.30} & 56.99\tiny{1.02} & 39.87\tiny{1.11} & 70.73\tiny{1.06} & 42.07\tiny{0.69} & 60.82\tiny{0.85} & 32.92\tiny{1.45} & 22.12\tiny{1.62} & 53.43\tiny{1.40} & 47.16\tiny{0.50}\\
    \midrule
    \multicolumn{6}{l}{\textcolor{black!50}{\textit{Architecture ablations}}}\\
    Hedgehog baseline                   & 26.45\tiny{1.29} & 52.74\tiny{1.02} & 38.38\tiny{1.10} & 68.01\tiny{1.09} & 30.60\tiny{0.64} & 54.80\tiny{0.87} & 30.43\tiny{1.42} & 21.66\tiny{1.62} & 50.91\tiny{1.41} & 40.79\tiny{0.49}\\
    +SSM                                & 26.62\tiny{1.29} & 52.27\tiny{1.02} & 38.74\tiny{1.10} & 69.21\tiny{1.08} & 32.95\tiny{0.65} & 53.39\tiny{0.87} & 29.67\tiny{1.41} & 22.43\tiny{1.64} & 50.83\tiny{1.41} & 41.07\tiny{0.49}\\
    \hspace{3pt}+Conv                   & 26.62\tiny{1.29} & 53.16\tiny{1.02} & 38.84\tiny{1.10} & 68.23\tiny{1.09} & 32.10\tiny{0.65} & 55.11\tiny{0.87} & 30.33\tiny{1.42} & 23.20\tiny{1.66} & 50.83\tiny{1.41} & 41.19\tiny{0.49}\\
    \band\hspace{6pt}+Gate(HedgeMamba)  & 26.96\tiny{1.30} & 53.83\tiny{1.02} & 38.59\tiny{1.10} & 69.21\tiny{1.08} & 32.68\tiny{0.65} & 55.47\tiny{0.87} & 30.53\tiny{1.43} & 22.73\tiny{1.64} & 51.78\tiny{1.40} & 41.70\tiny{0.49}\\
    \midrule
    \multicolumn{6}{l}{\textcolor{black!50}{\textit{Sensitivity: token allocation -- stage1 / stage2 (\%) split}}}\\
    100 / 0                             & 25.85\tiny{1.28} & 48.70\tiny{1.03} & 36.34\tiny{1.09} & 66.49\tiny{1.10} & 12.12\tiny{0.45} & 61.47\tiny{0.85} & 27.27\tiny{1.38} & 20.58\tiny{1.60} & 50.83\tiny{1.40} & 26.14\tiny{0.48}\\
    90 / 10                             & 25.34\tiny{1.27} & 52.53\tiny{1.02} & 38.59\tiny{1.10} & 68.66\tiny{1.08} & 29.69\tiny{0.64} & 53.52\tiny{0.87} & 30.14\tiny{1.42} & 21.97\tiny{1.62} & 51.62\tiny{1.40} & 39.93\tiny{0.49}\\
    75 / 25                             & 26.62\tiny{1.29} & 52.82\tiny{1.02} & 39.00\tiny{1.10} & 68.88\tiny{1.08} & 32.06\tiny{0.65} & 61.16\tiny{0.85} & 30.81\tiny{1.43} & 21.97\tiny{1.62} & 50.83\tiny{1.41} & 40.91\tiny{0.49}\\
    50 / 50                             & 26.96\tiny{1.30} & 53.83\tiny{1.02} & 38.59\tiny{1.10} & 69.21\tiny{1.08} & 32.68\tiny{0.65} & 55.47\tiny{0.87} & 30.53\tiny{1.43} & 22.73\tiny{1.64} & 51.78\tiny{1.40} & 41.70\tiny{0.49}\\
    25 / 75                             & 27.05\tiny{1.30} & 52.53\tiny{1.02} & 38.59\tiny{1.10} & 68.61\tiny{1.08} & 33.13\tiny{0.66} & 57.98\tiny{0.86} & 30.81\tiny{1.43} & 23.04\tiny{1.65} & 50.51\tiny{1.41} & 42.03\tiny{0.49}\\
    \band10 / 90                        & 27.22\tiny{1.30} & 53.24\tiny{1.02} & 40.23\tiny{1.11} & 68.72\tiny{1.08} & 32.74\tiny{0.65} & 55.38\tiny{0.87} & 30.62\tiny{1.43} & 23.20\tiny{1.66} & 51.85\tiny{1.40} & 42.38\tiny{0.49}\\
    0 / 100                             & 25.94\tiny{1.28} & 51.81\tiny{1.03} & 38.28\tiny{1.10} & 67.95\tiny{1.09} & 25.54\tiny{0.61} & 56.51\tiny{0.87} & 27.85\tiny{1.39} & 23.50\tiny{1.66} & 51.93\tiny{1.40} & 39.05\tiny{0.49}\\
    \midrule
    \multicolumn{6}{l}{\textcolor{black!50}{\textit{Scaling: overall token budget}}}\\
    1B                                  & 25.77\tiny{1.28} & 50.84\tiny{1.03} & 38.43\tiny{1.10} & 68.12\tiny{1.09} & 27.19\tiny{0.62} & 55.90\tiny{0.87} & 29.09\tiny{1.41} & 22.73\tiny{1.64} & 51.30\tiny{1.40} & 39.81\tiny{0.49}\\
    2B                                  & 26.02\tiny{1.28} & 51.43\tiny{1.03} & 38.54\tiny{1.10} & 69.21\tiny{1.08} & 29.73\tiny{0.64} & 58.53\tiny{0.86} & 29.67\tiny{1.41} & 21.81\tiny{1.62} & 50.99\tiny{1.40} & 40.54\tiny{0.49}\\
    3B                                  & 25.68\tiny{1.28} & 52.78\tiny{1.02} & 38.64\tiny{1.10} & 69.53\tiny{1.07} & 30.91\tiny{0.64} & 58.81\tiny{0.86} & 29.19\tiny{1.41} & 20.74\tiny{1.59} & 50.83\tiny{1.41} & 41.06\tiny{0.49}\\
    \band10B                            & 27.22\tiny{1.30} & 53.24\tiny{1.02} & 40.23\tiny{1.11} & 68.72\tiny{1.08} & 32.74\tiny{0.65} & 55.38\tiny{0.87} & 30.62\tiny{1.43} & 23.20\tiny{1.66} & 51.85\tiny{1.40} & 42.38\tiny{0.49}\\
    \bottomrule
\end{tabular}}
\end{table*}

In \cref{tab::scaling_baselines_exp} we report an additional analysis on the performance of our distillation procedure when applied to models of various sizes (160M and 410M, on top of our already-presented 1B results). Overall, the results confirms the trend of HedgeMamba consistently showing improvements over the Hedgehog approach.

\begin{table*}[t!]
\caption{Scaling analysis with respect to model size.}
\label{tab::scaling_baselines_exp}
\tiny
\centering
\resizebox{\textwidth}{!}{
\begin{tabular}{ll<{\hspace{-5pt}} c<{\hspace{-5pt}} *{10}{c<{\hspace{-7.5pt}}}}
    \toprule
    &Model & $\downarrow$~PPL  & $\uparrow$~Arc-C & Arc-E  & SIQA & PiQA
    &Lambada &BoolQ &RACE &LogiQA &WinoG & HSwag \\
    \midrule
    \multirow{2}{*}{\rotatebox{90}{160M}}
    & Pythia(Teacher)     & 39.38 & 23.63 & 43.64 & 36.75 & 62.30 & 22.38 & 56.88 & 28.71 & 19.05 & 51.22 & 30.28\\
    &Hedgehog (Baseline)  & 35.95 & 18.26 & 42.47 & 37.05 & 61.15 & 14.48 & 59.66 & 26.41 & 21.04 & 50.43 & 28.93\\
    &HedgeMamba (Ours)    & 24.18 & 23.81 & 44.57 & 36.44 & 61.32 & 16.03 & 55.44 & 25.65 & 21.04 & 50.43 & 29.96\\
    \midrule
    \multirow{2}{*}{\rotatebox{90}{410M}}
    & Pythia(Teacher)     & 16.50 & 24.32 & 51.89 & 38.95 & 66.70 & 36.60 & 60.58 & 30.72 & 21.97 & 53.27 & 40.62\\
    &Hedgehog (Baseline)  & 17.66 & 19.97 & 47.31 & 37.97 & 65.23 & 24.04 & 48.44 & 28.04 & 19.35 & 50.28 & 34.63\\
    &HedgeMamba (Ours)    & 16.05 & 24.06 & 48.86 & 39.00 & 66.00 & 27.85 & 52.32 & 27.56 & 20.58 & 52.17 & 36.64\\
    \midrule
    \multirow{2}{*}{\rotatebox{90}{1B}}
    &Pythia (Teacher)     & 13.86 & 27.05 & 56.99 & 39.87 & 70.73 & 42.07 & 60.82 & 32.92 & 22.12 & 53.43 & 47.16\\
    &Hedgehog (Baseline)  & 14.89 & 26.45 & 52.74 & 38.38 & 68.01 & 30.60 & 54.80 & 30.43 & 21.66 & 50.91 & 40.79\\
    &HedgeMamba (Ours)    & 14.01 & 27.22 & 53.24 & 40.23 & 68.72 & 32.74 & 55.38 & 30.62 & 23.20 & 51.85 & 42.38\\
    \bottomrule
\end{tabular}}
\end{table*}

\Cref{fig::plot_val_ppl} provides the detailed evolution of the validation perplexity during training, for the token allocation splits discussed in \cref{tab::token_budget_alloc}.
As a reminder, we train for a total 200K steps: each train step uses 49,200 tokens, so that the total number of tokens used in training are 49,200$\times$200,000=9,840,000,000 $\approx$ 10B.
The training steps are allocated between the first and second stage of our recipe in \cref{sec:recipe} depending on the split considered: for instance, a 10/90 split signifies that 10\% of the training steps (i.e. 20K steps) are used for stage 1 and 90\% (180K steps) for stage 2.
From \cref{fig::plot_val_ppl} we can infer that, even though stage 2 provides the larger benefits in decreasing overall PPL, the role of stage 1 is key in improving performance. Allocating all training tokens to stage 2, in fact, causes PPL to stall at a much higher value than if we allocated even a small fraction (with as small as 10\% giving the best results) also to stage 1.

\subsection{Implementation Details}
\label{app::implementation_details}
We use PyTorch with Distributed Data Parallel and mixed precision (bfloat16) for training. For our implementation of the HedgeMamba layer in \cref{fig::schematics_stages} we direclty adapt the Mamba-2 code, leveraging its chunk-wise parallel scan~\citep{dao2024transformers} (see the corresponding code in \cref{app::pseudocode}). With our default settings, stage 2 of distilling Pythia-1B on 10B tokens takes 29.8h (180K steps at 0.597 s/step) on a single 8$\times$A100-80GB GPU node.

We use the teacher models implementations and pretrained weights directly from the HuggingFace Transformers library~\citep{wolf-etal-2020-transformers}.
Student models are implemented by swapping the softmax Attention modules from the teacher with Mamba Mixer modules from~\cite{dao2024transformers}, equipped with the Hedgehog feature maps from~\cite{zhang2024hedgehog}. 

All the models are distilled on a compute node with 8 NVIDIA A100-80GB GPUs. We use AdamW optimizer ($\beta_1=0.9$, $\beta_2=0.95$) with linear warm-up and cosine decay to 0.1$\times$ peak LR schedule in our distillation procedure. Empirically, for models of size 1B, a peak learning rate of 0.01 was found to be suitable for stage 1, while a learning rate in the 1e-5 range worked best for stage 2. Following prior work~\citep{gu2023mamba, dao2024transformers}, we use gradient clipping of 1.0 and weight decay 0.1. 

\section{Additional Architecture details}
\label{app::additional_details}

\subsection{Full architectures schematics}

\begin{figure}[h]
    \centering
    \begin{minipage}{\linewidth}
        \centering
        \includegraphics[width=0.325\linewidth,trim={4cm 2cm 60cm 3cm},clip]{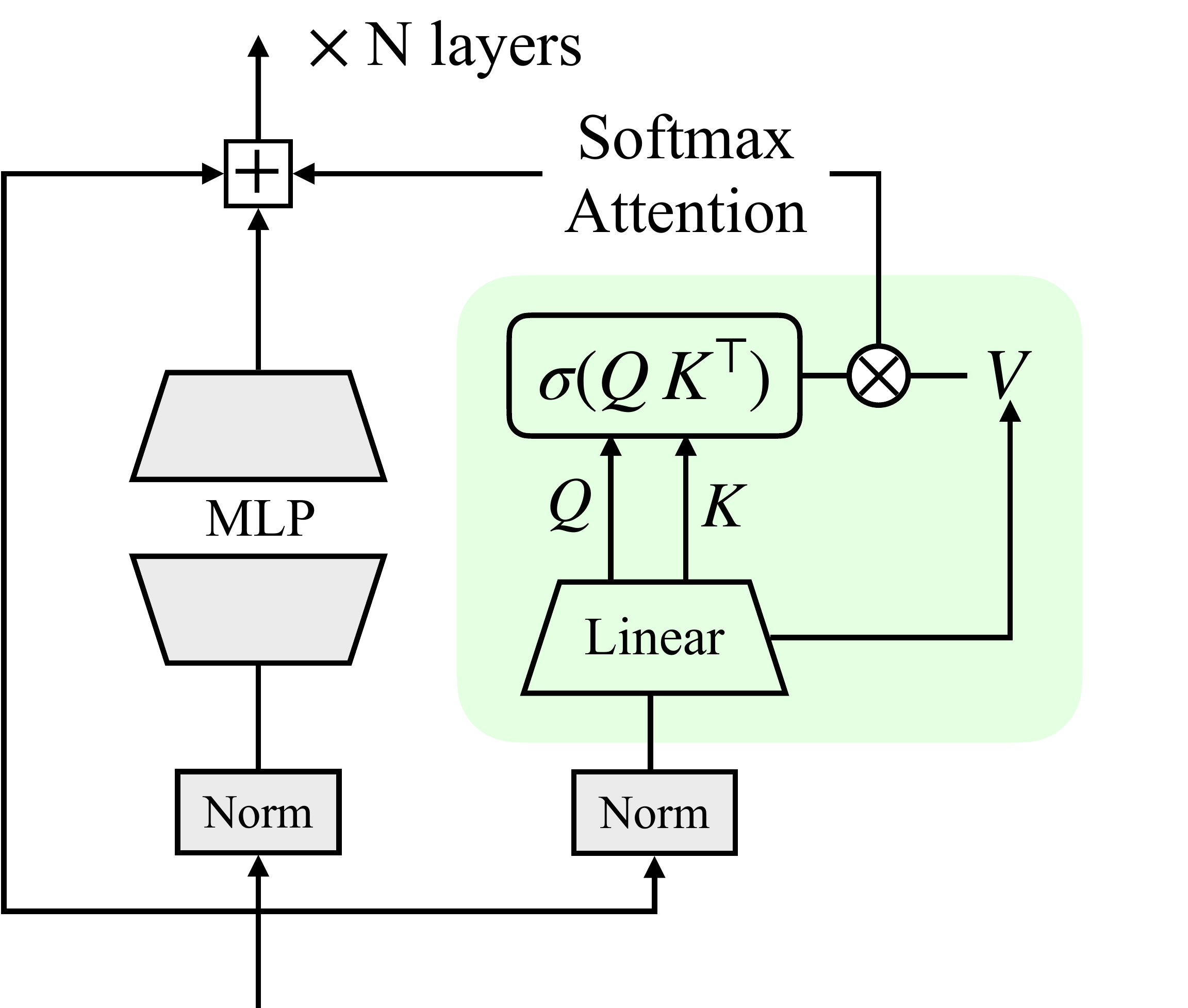}
        \par\vspace{1.5em}
    \end{minipage}
    \begin{minipage}[b]{0.325\linewidth} %
        \centering
        \includegraphics[width=\linewidth,trim={4cm 2cm 60cm 3cm},clip]{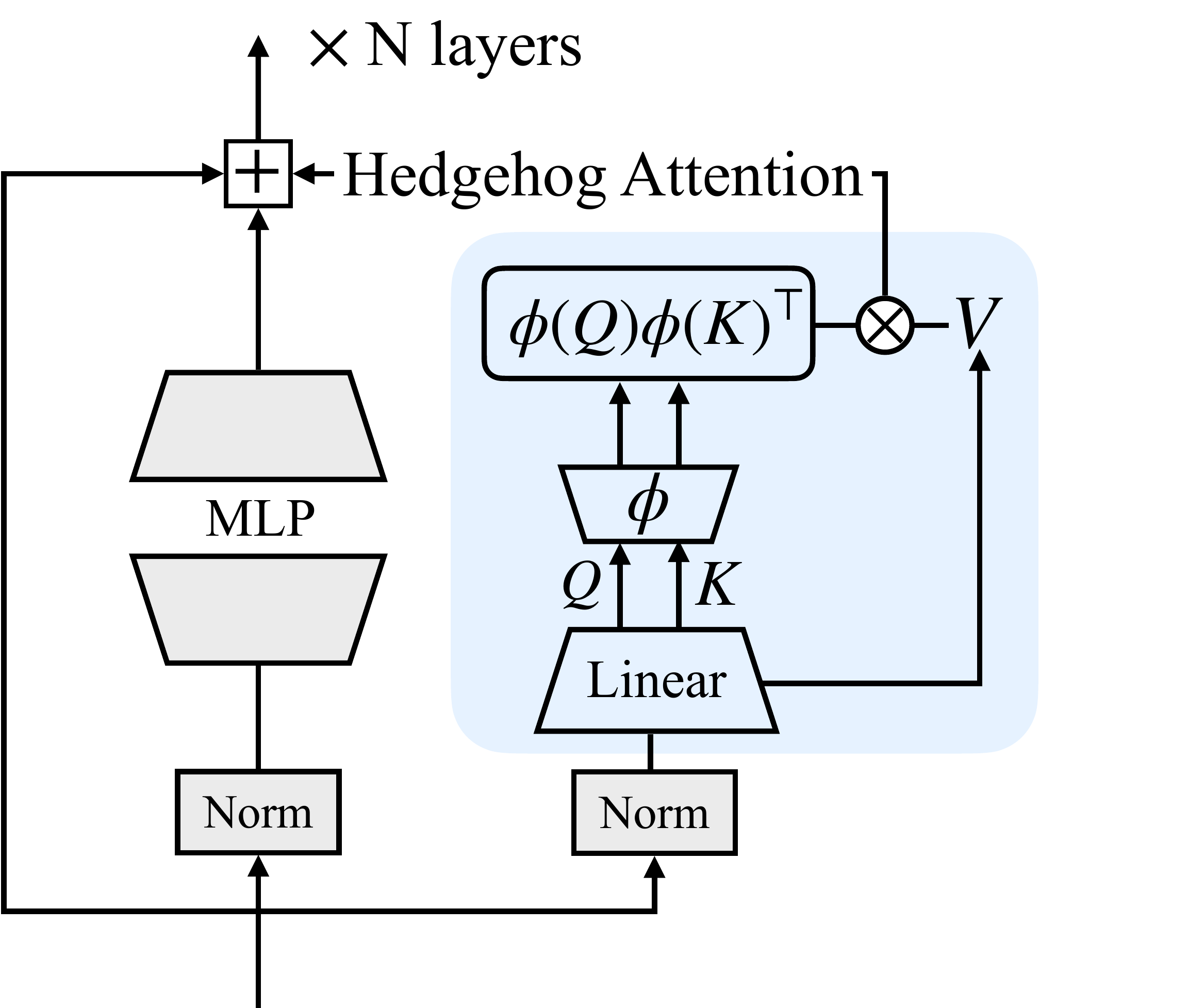}
    \end{minipage}
    \hfill
    \begin{minipage}[b]{0.325\linewidth} %
        \centering
        \includegraphics[width=\linewidth,trim={4cm 2cm 60cm 3cm},clip]{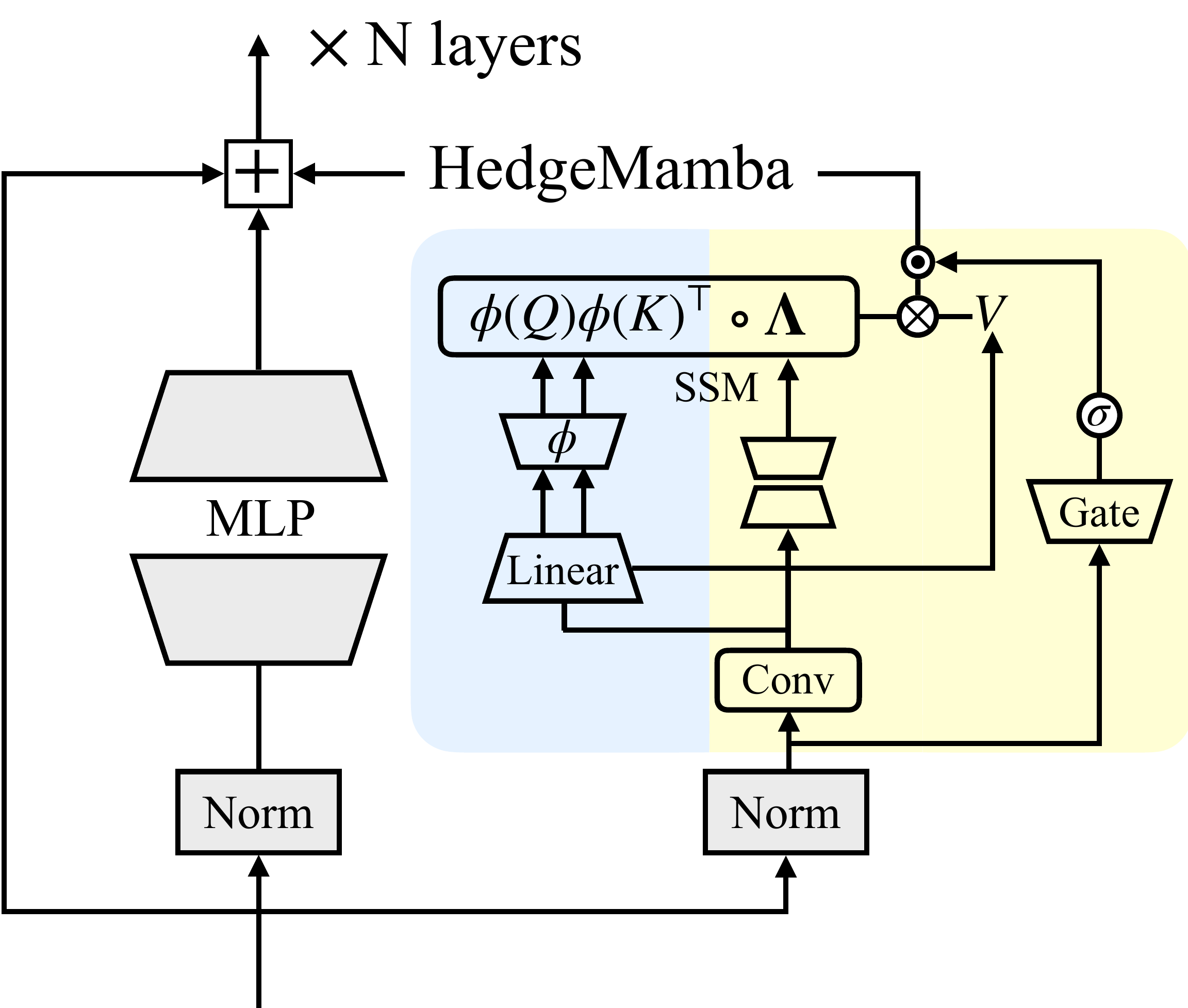}
    \end{minipage}
    \hfill
    \begin{minipage}[b]{0.325\linewidth} %
        \centering
        \includegraphics[width=\linewidth,trim={4cm 2cm 60cm 3cm},clip]{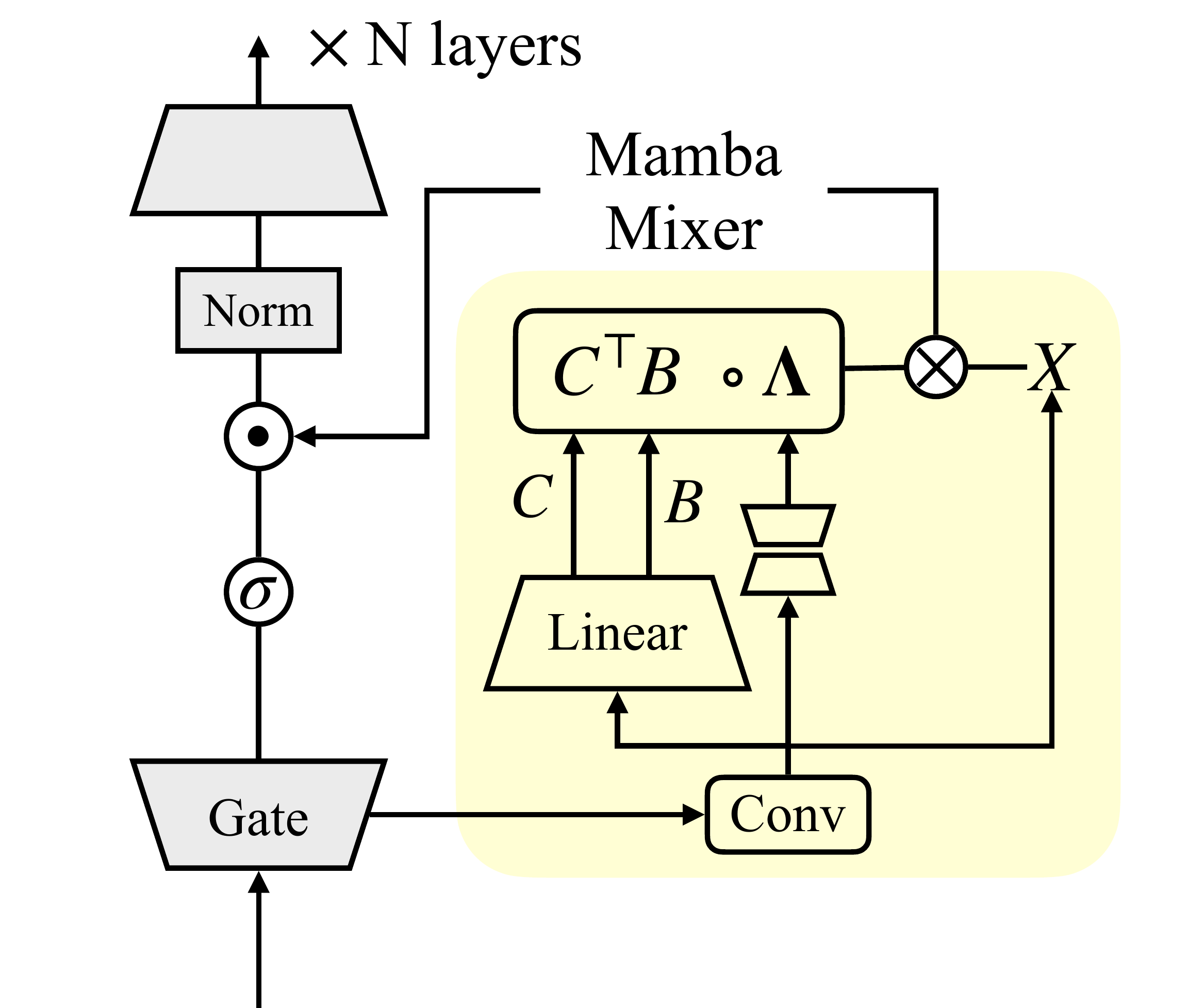}
    \end{minipage}
\caption{Schematics of architectures discussed in this project. Top: Pythia Transformer. Bottom, from left to right: Hedgehog, HedgeMamba, and Mamba.}
\label{fig::schematics}
\end{figure}

For reference, the diagrams in \cref{fig::schematics} describe the complete architectures discussed in this project. The Pythia Transformer \cite{biderman2023pythia}, which is the teacher model used throughout our experiments, appears on the top. The original Mamba architecture \cite{gu2023mamba} is reported on the bottom right. Notice how in both architectures the main components are a sequence mixer (Attention for the Transformer, and the SSM Mixer for Mamba) interwoven with MLPs (in Mamba, this role is covered by the gate branch). In the bottom left, we can see how the Hedgehog block attains Attention linearization. Finally, in the bottom-middle of \cref{fig::schematics}, acting as a bridge between the Hedgehog and Mamba architectures, we illustrate the HedgeMamba hybrid we proposed and used as student in this work: most of the architecture is inherited directly from Pythia, but the sequence mixer is substituted with a combination of Hedgehog and components from Mamba.

\subsection{Initialization of Mamba parameters in stage 2}
\label{app::hedgemamba_initialization}
A distinguishing feature of Mamba consists in its ability to prescribe a learnable causal mask \cite{dao2024transformers} via its state matrix $\bb{\Lambda}$ \cref{eqn:ssm_unrolled}. On top of this, the Mamba block also contains a short sequence-wise convolution and a gate branch. All these features contribute to set Mamba apart from other Linear Attention alternatives. We only leverage these three additional components in the second stage of our distillation method, while in the first stage we freeze them and make sure that they have no effect on the block output (see also the schematics in \cref{fig::schematics_stages}). This can be achieved by an opportune parameter initialization, described next.

\paragraph{State matrix}
The state matrix $\bb{\Lambda}$ in Mamba is obtained by exponentiating a product of two parameters: a rate-of-decay $\lambda\in\mathbb{R}^{N\times d}$ and a time-step $\Delta=\Delta(\bb{X})\in\mathbb{R}^{N}$. In particular, the latter is recovered by applying an MLP to the input $\bb{X}\in\mathbb{R}^{L\times d}$: two linear applications with weights and biases $(\bb{W}_d,\bb{b}_d)\in\mathbb{R}^{d\times d_r}\times\mathbb{R}^{d_r}$ and $(\bb{W}_u,\bb{b}_u)\in\mathbb{R}^{d_r\times N}\times\mathbb{R}^{N}$, respectively, followed by a \texttt{SoftPlus} nonlinearity. The overall formula is given by
\begin{equation}
    \bb{\Lambda}_l = e^{-\lambda\odot(\Delta_l\otimes\bb{1}^\top)},
    \quad\text{with}\quad
    \Delta_l=\text{\texttt{SoftPlus}}((\bb{X}_{l,:}\bb{W}_d+\bb{b}_d)\bb{W}_u+\bb{b}_u),
    \qquad\forall l=1\dots L.
\end{equation}
To reduce $\bb{\Lambda}_l$ to identity at initialization, we would need to impose $\lambda\equiv0$. However, Mamba parameterizes the decay rate $\lambda$ in log-space, meaning we would have to drive $\log\lambda\to-\infty$. This collapses gradient flow, implying that with this initialization, $\lambda$ would struggle to move away from $0$. We therefore introduce a separate multiplicative gate $s$, clamped in $[0,1]$, and consider $\lambda_{\text{eff}} = s\lambda$, setting $s=0$ at initialization. With this, gradient can still flow through both $s$ and $\lambda$, and the gate can freely learn and move away from identity.
Finally, note that $\Delta_l$ also affects the definition of $\bb{B}$ in \cref{eqn:ssm_unrolled}: to nullify its effect, we must have $\Delta_l\equiv\bb{1}$. To this end, we also impose $\bb{W}_u\equiv\bb{0}$ and $\bb{b}_u\equiv\text{\texttt{SoftPlus}}^{-1}(1)\cdot\bb{1}\approx 0.541324\cdot\bb{1}$. Despite this initialization, gradient can flow freely through $\Delta_l$ from step 1, so no adjustments are needed here.

\paragraph{Convolution}
The convolution component applies the following operation: given an input $\bb{X}\in\mathbb{R}^{L\times d}$, and its kernel weights $\bb{W}\in\mathbb{R}^{\kappa\times d}$ (for a kernel of size $\kappa$) and biases $\bb{b}\in\mathbb{R}^d$, the output is given by:
\begin{equation}
    \bb{Y}_{l,:} = \bb{b} + \sum_{i=1}^{\kappa} \bb{W}_{i,:}\odot\bb{X}_{l-\kappa+i,:} \, .
\end{equation}
To collapse this to the identity operation, it suffices then to pick $\bb{b}\equiv\bb{0}$, $\bb{W}_{\kappa,:}\equiv\bb{1}$, and $\bb{W}_{i\neq\kappa,:}\equiv\bb{0}$. Notice that in the original Mamba block, the convolution is followed by a nonlinearity, which acts before the SSM mixing layer. In our architecture, we relocate this nonlinearity after the query and key projections (within the Hedgehog feature map \cref{eqn:hedgehog_MLP}): this is necessary for HedgeMamba to exactly reproduce, at initialization, the Linear Attention solution learned in stage 1.

\paragraph{Gate}
The gate branch, on the other hand, consists of a linear layer (of weights $\bb{W}\in\mathbb{R}^{d\times d}$ and biases $\bb{b}\in\mathbb{R}^d$), followed by a \texttt{SiLU} nonlinearity. The output is then element-wise multiplied by the output of the SSM mixer (here denoted as $\bb{X}_{SSM}$). Overall, this amounts to
\begin{equation}
    \bb{Y} = \bb{X}_{SSM}\odot\text{\texttt{SiLU}}(\bb{X}\bb{W}+\bb{b}).
\end{equation}
To obtain the identity, then, it suffices to set $\bb{W}\equiv\bb{0}$, and $\bb{b}=\text{\texttt{SiLU}}^{-1}(1)\cdot\bb{1}\approx 1.27846\cdot\bb{1}$.

\newpage
\section{Pseudocode}
\label{app::pseudocode}
Here we provide pseudocode for the application of the forward pass of the student model used in our experiments.

More in detail, \cref{code::pymamba_layer} reports the implementation of the whole adapted Pythia block \citep{biderman2023pythia}. As in the original Pythia model, the flow of operations in this block is split into two branches (see also \cref{fig::schematics}). On the one hand, we have an MLP with pre-normalization; on the other, the vanilla Attention layer is substituted with the hybrid HedgeMamba layer described in \cref{sec:recipe:stage2}.

\begin{listing}[b!]
\begin{Verbatim}[commandchars=\\\{\},fontsize=\small]
\PYG{c+c1}{\PYGZsh{} \PYGZhy{}\PYGZhy{}\PYGZhy{} modified pythia layer powered by hedge\PYGZhy{}mamba module \PYGZhy{}\PYGZhy{}\PYGZhy{} \PYGZsh{}}
\PYG{k}{class}\PYG{+w}{ }\PYG{n+nc}{HedgeMambaLayer}\PYG{p}{(}\PYG{n}{GPTNeoXLayer}\PYG{p}{)}\PYG{p}{:}
    \PYG{k}{def}\PYG{+w}{ }\PYG{n+nf+fm}{\PYGZus{}\PYGZus{}init\PYGZus{}\PYGZus{}}\PYG{p}{(}
        \PYG{n+nb+bp}{self}\PYG{p}{,}
        \PYG{n}{config}\PYG{p}{:} \PYG{n}{PretrainedConfig}\PYG{p}{,}
    \PYG{p}{)}\PYG{p}{:}
        \PYG{n+nb}{super}\PYG{p}{(}\PYG{p}{)}\PYG{o}{.}\PYG{n+nf+fm}{\PYGZus{}\PYGZus{}init\PYGZus{}\PYGZus{}}\PYG{p}{(}\PYG{n}{config}\PYG{p}{)}  \PYG{c+c1}{\PYGZsh{} standard pythia layer init}
        \PYG{n+nb+bp}{self}\PYG{o}{.}\PYG{n}{mixer\PYGZus{}config} \PYG{o}{=} \PYG{n}{mixer\PYGZus{}config}
        \PYG{c+c1}{\PYGZsh{} overwrite attention module with mamba\PYGZhy{}mixer}
        \PYG{n+nb+bp}{self}\PYG{o}{.}\PYG{n}{attention} \PYG{o}{=} \PYG{n}{HedgeMambaMixer}\PYG{p}{(}\PYG{n}{config}\PYG{p}{)}

    \PYG{k}{def}\PYG{+w}{ }\PYG{n+nf}{forward}\PYG{p}{(}
        \PYG{n+nb+bp}{self}\PYG{p}{,}
        \PYG{n}{hidden\PYGZus{}states}\PYG{p}{:} \PYG{n}{torch}\PYG{o}{.}\PYG{n}{FloatTensor}\PYG{p}{,}
        \PYG{n}{cache\PYGZus{}params}\PYG{p}{:} \PYG{n}{Optional}\PYG{p}{[}\PYG{n}{MambaCache}\PYG{p}{]} \PYG{o}{=} \PYG{k+kc}{None}\PYG{p}{,}
    \PYG{p}{)}\PYG{p}{:}
        \PYG{c+c1}{\PYGZsh{} attention stream}
        \PYG{n}{attn\PYGZus{}output} \PYG{o}{=} \PYG{n+nb+bp}{self}\PYG{o}{.}\PYG{n}{input\PYGZus{}layernorm}\PYG{p}{(}\PYG{n}{hidden\PYGZus{}states}\PYG{p}{)}
        \PYG{n}{attn\PYGZus{}output} \PYG{o}{=} \PYG{n+nb+bp}{self}\PYG{o}{.}\PYG{n}{attention}\PYG{p}{(}\PYG{n}{attn\PYGZus{}output}\PYG{p}{,} \PYG{n}{cache\PYGZus{}params}\PYG{o}{=}\PYG{n}{cache\PYGZus{}params}\PYG{p}{)}
        \PYG{n}{attn\PYGZus{}output} \PYG{o}{=} \PYG{n+nb+bp}{self}\PYG{o}{.}\PYG{n}{post\PYGZus{}attention\PYGZus{}dropout}\PYG{p}{(}\PYG{n}{attn\PYGZus{}output}\PYG{p}{)}

        \PYG{c+c1}{\PYGZsh{} mlp stream}
        \PYG{n}{mlp\PYGZus{}output} \PYG{o}{=} \PYG{n+nb+bp}{self}\PYG{o}{.}\PYG{n}{mlp}\PYG{p}{(}\PYG{n+nb+bp}{self}\PYG{o}{.}\PYG{n}{post\PYGZus{}attention\PYGZus{}layernorm}\PYG{p}{(}\PYG{n}{hidden\PYGZus{}states}\PYG{p}{)}\PYG{p}{)}
        \PYG{n}{mlp\PYGZus{}output} \PYG{o}{=} \PYG{n+nb+bp}{self}\PYG{o}{.}\PYG{n}{post\PYGZus{}mlp\PYGZus{}dropout}\PYG{p}{(}\PYG{n}{mlp\PYGZus{}output}\PYG{p}{)}

        \PYG{c+c1}{\PYGZsh{} pythia layer with parallel MLP and attention streams}
        \PYG{c+c1}{\PYGZsh{} pseudocode: x = x + attn(ln1(x)) + mlp(ln2(x))}
        \PYG{n}{hidden\PYGZus{}states} \PYG{o}{=} \PYG{n}{hidden\PYGZus{}states} \PYG{o}{+} \PYG{n}{attn\PYGZus{}output} \PYG{o}{+} \PYG{n}{mlp\PYGZus{}output}
        \PYG{k}{return} \PYG{n}{hidden\PYGZus{}states}
\end{Verbatim}
\caption{Implementation of our \texttt{PyMambaLayer}, which replaces the Softmax Attention module in the Pythia Transformer with our \texttt{HedgeMamba} mixer. Code for the latter is provided in \cref{code::hedgemamba}.}
\label{code::pymamba_layer}
\end{listing}

HedgeMamba is the core module introduced in our work, and pseudocode for its implementation is detailed in \cref{code::hedgemamba}. Its code blueprint closely follows the one for the Mamba SSM mixer \citep{dao2024transformers}, including a gate branch and a short convolution before the mixer application, but presents three main differences: (i) the SSM parameters $\bb{B},\bb{C}$ (covering the roles of \emph{keys} and \emph{queries} in Linear Attention) are further modified according to the Hedgehog map; (ii) the input to the SSM is mapped through an additional linear layer to recover the \emph{values} in the linearized version of Attention; (iii) the SSM hidden state is expanded to accommodate for normalization terms, as per \cref{eqn:state_expansion}.

\begin{listing}[t!]
\begin{Verbatim}[commandchars=\\\{\},fontsize=\small]
\PYG{c+c1}{\PYGZsh{} \PYGZhy{}\PYGZhy{}\PYGZhy{} hedge\PYGZhy{}mamba mixer module \PYGZhy{}\PYGZhy{}\PYGZhy{} \PYGZsh{}}
\PYG{k}{class}\PYG{+w}{ }\PYG{n+nc}{HedgeMambaMixer}\PYG{p}{(}\PYG{n}{nn}\PYG{o}{.}\PYG{n}{Module}\PYG{p}{)}\PYG{p}{:}
    \PYG{k}{def}\PYG{+w}{ }\PYG{n+nf+fm}{\PYGZus{}\PYGZus{}init\PYGZus{}\PYGZus{}}\PYG{p}{(}\PYG{n+nb+bp}{self}\PYG{p}{,} \PYG{n}{config}\PYG{p}{)}\PYG{p}{:}
        \PYG{n+nb}{super}\PYG{p}{(}\PYG{p}{)}\PYG{o}{.}\PYG{n+nf+fm}{\PYGZus{}\PYGZus{}init\PYGZus{}\PYGZus{}}\PYG{p}{(}\PYG{p}{)}
        \PYG{n+nb+bp}{self}\PYG{o}{.}\PYG{n}{hidden\PYGZus{}size\PYGZus{}per\PYGZus{}head} \PYG{o}{=} \PYG{n}{config}\PYG{o}{.}\PYG{n}{hidden\PYGZus{}size} \PYG{o}{/}\PYG{o}{/} \PYG{n}{config}\PYG{o}{.}\PYG{n}{num\PYGZus{}attention\PYGZus{}heads}

        \PYG{c+c1}{\PYGZsh{} from mamba}
        \PYG{c+c1}{\PYGZsh{} state\PYGZus{}size == hidden\PYGZus{}size to mimic attention}
        \PYG{n+nb+bp}{self}\PYG{o}{.}\PYG{n}{gate\PYGZus{}proj} \PYG{o}{=} \PYG{n}{nn}\PYG{o}{.}\PYG{n}{Linear}\PYG{p}{(}\PYG{n}{config}\PYG{o}{.}\PYG{n}{hidden\PYGZus{}size}\PYG{p}{,} \PYG{n}{config}\PYG{o}{.}\PYG{n}{hidden\PYGZus{}size}\PYG{p}{)}
        \PYG{n+nb+bp}{self}\PYG{o}{.}\PYG{n}{conv1d} \PYG{o}{=} \PYG{n}{nn}\PYG{o}{.}\PYG{n}{Conv1d}\PYG{p}{(}\PYG{n}{config}\PYG{o}{.}\PYG{n}{hidden\PYGZus{}size}\PYG{p}{,} \PYG{n}{hidden\PYGZus{}size}\PYG{p}{)}
        \PYG{n+nb+bp}{self}\PYG{o}{.}\PYG{n}{x\PYGZus{}proj} \PYG{o}{=} \PYG{n}{nn}\PYG{o}{.}\PYG{n}{Linear}\PYG{p}{(}
            \PYG{n}{config}\PYG{o}{.}\PYG{n}{hidden\PYGZus{}size}\PYG{p}{,} \PYG{n}{config}\PYG{o}{.}\PYG{n}{time\PYGZus{}step\PYGZus{}rank} \PYG{o}{+} \PYG{n}{config}\PYG{o}{.}\PYG{n}{hidden\PYGZus{}size} \PYG{o}{*} \PYG{l+m+mi}{2}
        \PYG{p}{)}
        \PYG{n+nb+bp}{self}\PYG{o}{.}\PYG{n}{A\PYGZus{}log} \PYG{o}{=} \PYG{n}{nn}\PYG{o}{.}\PYG{n}{Parameter}\PYG{p}{(}\PYG{n+nb+bp}{self}\PYG{o}{.}\PYG{n}{init\PYGZus{}A}\PYG{p}{(}\PYG{n}{config}\PYG{p}{)}\PYG{p}{)}  \PYG{c+c1}{\PYGZsh{} decay, lambda = exp(A\PYGZus{}log)}
        \PYG{n+nb+bp}{self}\PYG{o}{.}\PYG{n}{lambda\PYGZus{}gate} \PYG{o}{=} \PYG{n}{nn}\PYG{o}{.}\PYG{n}{Parameter}\PYG{p}{(}\PYG{n}{torch}\PYG{o}{.}\PYG{n}{zeros}\PYG{p}{(}\PYG{n}{num\PYGZus{}heads}\PYG{p}{)}\PYG{p}{)}  \PYG{c+c1}{\PYGZsh{} s in [0,1], init 0}
        \PYG{n+nb+bp}{self}\PYG{o}{.}\PYG{n}{dt\PYGZus{}proj} \PYG{o}{=} \PYG{n}{nn}\PYG{o}{.}\PYG{n}{Linear}\PYG{p}{(}\PYG{n}{config}\PYG{o}{.}\PYG{n}{time\PYGZus{}step\PYGZus{}rank}\PYG{p}{,} \PYG{n}{num\PYGZus{}heads}\PYG{p}{)}
        \PYG{n+nb+bp}{self}\PYG{o}{.}\PYG{n}{out\PYGZus{}proj} \PYG{o}{=} \PYG{n}{nn}\PYG{o}{.}\PYG{n}{Linear}\PYG{p}{(}\PYG{n}{config}\PYG{o}{.}\PYG{n}{hidden\PYGZus{}size}\PYG{p}{,} \PYG{n}{config}\PYG{o}{.}\PYG{n}{hidden\PYGZus{}size}\PYG{p}{)}

        \PYG{c+c1}{\PYGZsh{} additional projections to replicate linear attention}
        \PYG{n+nb+bp}{self}\PYG{o}{.}\PYG{n}{v\PYGZus{}proj} \PYG{o}{=} \PYG{n}{nn}\PYG{o}{.}\PYG{n}{Linear}\PYG{p}{(}\PYG{n}{config}\PYG{o}{.}\PYG{n}{hidden\PYGZus{}size}\PYG{p}{,} \PYG{n}{config}\PYG{o}{.}\PYG{n}{hidden\PYGZus{}size}\PYG{p}{)}  \PYG{c+c1}{\PYGZsh{} values}
        \PYG{n+nb+bp}{self}\PYG{o}{.}\PYG{n}{hhog\PYGZus{}q} \PYG{o}{=} \PYG{n}{HedgehogProjection}\PYG{p}{(}\PYG{n}{config}\PYG{p}{,} \PYG{n+nb+bp}{self}\PYG{o}{.}\PYG{n}{hidden\PYGZus{}size\PYGZus{}per\PYGZus{}head}\PYG{p}{)}  \PYG{c+c1}{\PYGZsh{} hedgehog}
        \PYG{n+nb+bp}{self}\PYG{o}{.}\PYG{n}{hhog\PYGZus{}k} \PYG{o}{=} \PYG{n}{HedgehogProjection}\PYG{p}{(}\PYG{n}{config}\PYG{p}{,} \PYG{n+nb+bp}{self}\PYG{o}{.}\PYG{n}{hidden\PYGZus{}size\PYGZus{}per\PYGZus{}head}\PYG{p}{)}
        \PYG{n+nb+bp}{self}\PYG{o}{.}\PYG{n}{rotary\PYGZus{}ndims} \PYG{o}{=} \PYG{n+nb}{int}\PYG{p}{(}\PYG{n+nb+bp}{self}\PYG{o}{.}\PYG{n}{hidden\PYGZus{}size\PYGZus{}per\PYGZus{}head} \PYG{o}{*} \PYG{n}{config}\PYG{o}{.}\PYG{n}{rotary\PYGZus{}pct}\PYG{p}{)}
        \PYG{n+nb+bp}{self}\PYG{o}{.}\PYG{n}{\PYGZus{}init\PYGZus{}rope}\PYG{p}{(}\PYG{p}{)}  \PYG{c+c1}{\PYGZsh{} rope positional encoding}

    \PYG{k}{def}\PYG{+w}{ }\PYG{n+nf}{forward}\PYG{p}{(}\PYG{n+nb+bp}{self}\PYG{p}{,} \PYG{n}{hidden\PYGZus{}states}\PYG{p}{:} \PYG{n}{torch}\PYG{o}{.}\PYG{n}{Tensor}\PYG{p}{)} \PYG{o}{\PYGZhy{}}\PYG{o}{\PYGZgt{}} \PYG{n}{torch}\PYG{o}{.}\PYG{n}{Tensor}\PYG{p}{:}
        \PYG{n}{gate} \PYG{o}{=} \PYG{n}{F}\PYG{o}{.}\PYG{n}{silu}\PYG{p}{(}\PYG{n+nb+bp}{self}\PYG{o}{.}\PYG{n}{gate\PYGZus{}proj}\PYG{p}{(}\PYG{n}{hidden\PYGZus{}states}\PYG{p}{)}\PYG{p}{)}
        \PYG{n}{hidden\PYGZus{}states} \PYG{o}{=} \PYG{n+nb+bp}{self}\PYG{o}{.}\PYG{n}{conv1d}\PYG{p}{(}\PYG{n}{hidden\PYGZus{}states}\PYG{p}{)}

        \PYG{c+c1}{\PYGZsh{} linear proj to recover SSM parameters}
        \PYG{n}{dt}\PYG{p}{,} \PYG{n}{B}\PYG{p}{,} \PYG{n}{C} \PYG{o}{=} \PYG{n}{torch}\PYG{o}{.}\PYG{n}{split}\PYG{p}{(}\PYG{n+nb+bp}{self}\PYG{o}{.}\PYG{n}{x\PYGZus{}proj}\PYG{p}{(}\PYG{n}{hidden\PYGZus{}states}\PYG{p}{)}\PYG{p}{,}
                                \PYG{p}{[}\PYG{n}{time\PYGZus{}rank}\PYG{p}{,} \PYG{n}{state\PYGZus{}size}\PYG{p}{,} \PYG{n}{state\PYGZus{}size}\PYG{p}{]}\PYG{p}{,} \PYG{n}{dim}\PYG{o}{=}\PYG{o}{\PYGZhy{}}\PYG{l+m+mi}{1}\PYG{p}{)}

        \PYG{c+c1}{\PYGZsh{} rope positional encoding as in pythia}
        \PYG{n}{B} \PYG{o}{=} \PYG{n}{B}\PYG{o}{.}\PYG{n}{view}\PYG{p}{(}\PYG{n}{batch\PYGZus{}size}\PYG{p}{,} \PYG{n}{seq\PYGZus{}len}\PYG{p}{,} \PYG{n}{num\PYGZus{}heads}\PYG{p}{,} \PYG{n}{hidden\PYGZus{}size\PYGZus{}per\PYGZus{}head}\PYG{p}{)}
        \PYG{n}{C} \PYG{o}{=} \PYG{n}{C}\PYG{o}{.}\PYG{n}{view}\PYG{p}{(}\PYG{n}{batch\PYGZus{}size}\PYG{p}{,} \PYG{n}{seq\PYGZus{}len}\PYG{p}{,} \PYG{n}{num\PYGZus{}heads}\PYG{p}{,} \PYG{n}{hidden\PYGZus{}size\PYGZus{}per\PYGZus{}head}\PYG{p}{)}
        \PYG{n}{C} \PYG{o}{=} \PYG{n+nb+bp}{self}\PYG{o}{.}\PYG{n}{rotary\PYGZus{}emb}\PYG{p}{(}\PYG{n}{C}\PYG{p}{,} \PYG{n}{seq\PYGZus{}len}\PYG{o}{=}\PYG{n}{C}\PYG{o}{.}\PYG{n}{shape}\PYG{p}{[}\PYG{l+m+mi}{1}\PYG{p}{]}\PYG{p}{)}  \PYG{c+c1}{\PYGZsh{} equivalent to Q from attention}
        \PYG{n}{B} \PYG{o}{=} \PYG{n+nb+bp}{self}\PYG{o}{.}\PYG{n}{rotary\PYGZus{}emb}\PYG{p}{(}\PYG{n}{B}\PYG{p}{,} \PYG{n}{seq\PYGZus{}len}\PYG{o}{=}\PYG{n}{B}\PYG{o}{.}\PYG{n}{shape}\PYG{p}{[}\PYG{l+m+mi}{1}\PYG{p}{]}\PYG{p}{)}  \PYG{c+c1}{\PYGZsh{} equivalent to K from attention}

        \PYG{c+c1}{\PYGZsh{} apply hedgehog feature map}
        \PYG{n}{B}\PYG{p}{,} \PYG{n}{C} \PYG{o}{=} \PYG{n+nb+bp}{self}\PYG{o}{.}\PYG{n}{hhog\PYGZus{}k}\PYG{p}{(}\PYG{n}{B}\PYG{p}{)}\PYG{p}{,} \PYG{n+nb+bp}{self}\PYG{o}{.}\PYG{n}{hhog\PYGZus{}q}\PYG{p}{(}\PYG{n}{C}\PYG{p}{)}

        \PYG{c+c1}{\PYGZsh{} decay rate gated by s: s = 0 at init, so Lambda == 1 for any dt and the}
        \PYG{c+c1}{\PYGZsh{} block starts exactly at the stage\PYGZhy{}1 linear attention solution}
        \PYG{n}{A} \PYG{o}{=} \PYG{o}{\PYGZhy{}}\PYG{n}{torch}\PYG{o}{.}\PYG{n}{exp}\PYG{p}{(}\PYG{n+nb+bp}{self}\PYG{o}{.}\PYG{n}{A\PYGZus{}log}\PYG{p}{)} \PYG{o}{*} \PYG{n+nb+bp}{self}\PYG{o}{.}\PYG{n}{lambda\PYGZus{}gate}\PYG{o}{.}\PYG{n}{clamp}\PYG{p}{(}\PYG{l+m+mf}{0.0}\PYG{p}{,} \PYG{l+m+mf}{1.0}\PYG{p}{)}

        \PYG{c+c1}{\PYGZsh{} value projection + expand V with 1 for score normalization as in attention}
        \PYG{c+c1}{\PYGZsh{} decay is one scalar per head, so a single extra channel suffices}
        \PYG{n}{V} \PYG{o}{=} \PYG{n+nb+bp}{self}\PYG{o}{.}\PYG{n}{v\PYGZus{}proj}\PYG{p}{(}\PYG{n}{hidden\PYGZus{}states}\PYG{p}{)}
        \PYG{n}{V} \PYG{o}{=} \PYG{n}{torch}\PYG{o}{.}\PYG{n}{cat}\PYG{p}{(}\PYG{p}{[}\PYG{n}{V}\PYG{p}{,} \PYG{n}{torch}\PYG{o}{.}\PYG{n}{ones\PYGZus{}like}\PYG{p}{(}\PYG{n}{V}\PYG{p}{[}\PYG{o}{.}\PYG{o}{.}\PYG{o}{.}\PYG{p}{,} \PYG{p}{:}\PYG{l+m+mi}{1}\PYG{p}{]}\PYG{p}{)}\PYG{p}{]}\PYG{p}{,} \PYG{n}{dim}\PYG{o}{=}\PYG{o}{\PYGZhy{}}\PYG{l+m+mi}{1}\PYG{p}{)}
        \PYG{n}{scan\PYGZus{}outputs} \PYG{o}{=} \PYG{n}{mamba\PYGZus{}chunk\PYGZus{}scan\PYGZus{}combined}\PYG{p}{(}\PYG{n}{V}\PYG{p}{,} \PYG{n}{dt}\PYG{p}{,} \PYG{n}{A}\PYG{p}{,} \PYG{n}{B}\PYG{p}{,} \PYG{n}{C}\PYG{p}{)}  \PYG{c+c1}{\PYGZsh{} Mamba 2 SSD mixer}
        \PYG{n}{scan\PYGZus{}outputs} \PYG{o}{=} \PYG{n}{gate} \PYG{o}{*} \PYG{n}{scan\PYGZus{}outputs}\PYG{p}{[}\PYG{o}{.}\PYG{o}{.}\PYG{o}{.}\PYG{p}{,} \PYG{p}{:}\PYG{o}{\PYGZhy{}}\PYG{l+m+mi}{1}\PYG{p}{]} \PYG{o}{/} \PYG{n}{scan\PYGZus{}outputs}\PYG{p}{[}\PYG{o}{.}\PYG{o}{.}\PYG{o}{.}\PYG{p}{,} \PYG{o}{\PYGZhy{}}\PYG{l+m+mi}{1}\PYG{p}{:}\PYG{p}{]}

        \PYG{k}{return} \PYG{n+nb+bp}{self}\PYG{o}{.}\PYG{n}{out\PYGZus{}proj}\PYG{p}{(}\PYG{n}{scan\PYGZus{}outputs}\PYG{p}{)}
\end{Verbatim}
\caption{PyTorch-style pseudocode of our \texttt{HedgeMamba} sequence mixer, which equips the vanilla Mamba SSM mixer with the Hedgehog feature map~\citep{zhang2024hedgehog} for Attention linearization. The gate \texttt{lambda\_gate} is the multiplicative $s$ of \cref{app::hedgemamba_initialization}, which keeps $\bb{\Lambda}\equiv\mI$ at initialization without driving $\log\lambda\to-\infty$.}
\label{code::hedgemamba}
\end{listing}

\FloatBarrier
Finally, in \cref{code::hedgehog} we report also our implementation of the Hedgehog projection operator, used within the HedgeMamba layer. Like in the original Hedgehog paper, the output of the feature map $\bb{\phi}$ \cref{eqn:hedgehog_MLP}, is duplicated by collating its opposite. As a nonlinearity, we apply a softmax operation \emph{along the embedding dimension}, instead of vanilla exponentiation as done in \citet{zhang2024hedgehog}: this is to guarantee better numerical stability.

\begin{listing}[h!]
\begin{Verbatim}[commandchars=\\\{\},fontsize=\small]
\PYG{c+c1}{\PYGZsh{} \PYGZhy{}\PYGZhy{}\PYGZhy{} hedgehog projection module \PYGZhy{}\PYGZhy{}\PYGZhy{} \PYGZsh{}}
\PYG{k}{class}\PYG{+w}{ }\PYG{n+nc}{HedgehogProjection}\PYG{p}{(}\PYG{n}{nn}\PYG{o}{.}\PYG{n}{Module}\PYG{p}{)}\PYG{p}{:}
    \PYG{k}{def}\PYG{+w}{ }\PYG{n+nf+fm}{\PYGZus{}\PYGZus{}init\PYGZus{}\PYGZus{}}\PYG{p}{(}\PYG{n+nb+bp}{self}\PYG{p}{,} \PYG{n}{config}\PYG{p}{,} \PYG{n}{head\PYGZus{}size}\PYG{p}{,} \PYG{n}{bias}\PYG{o}{=}\PYG{k+kc}{True}\PYG{p}{)}\PYG{p}{:}
        \PYG{n+nb}{super}\PYG{p}{(}\PYG{p}{)}\PYG{o}{.}\PYG{n+nf+fm}{\PYGZus{}\PYGZus{}init\PYGZus{}\PYGZus{}}\PYG{p}{(}\PYG{p}{)}
        \PYG{n+nb+bp}{self}\PYG{o}{.}\PYG{n}{config} \PYG{o}{=} \PYG{n}{config}
        \PYG{n+nb+bp}{self}\PYG{o}{.}\PYG{n}{phi} \PYG{o}{=} \PYG{n}{nn}\PYG{o}{.}\PYG{n}{Linear}\PYG{p}{(}\PYG{n}{head\PYGZus{}size}\PYG{p}{,} \PYG{n}{head\PYGZus{}size}\PYG{p}{,} \PYG{n}{bias}\PYG{o}{=}\PYG{n}{bias}\PYG{p}{)}

    \PYG{k}{def}\PYG{+w}{ }\PYG{n+nf}{forward}\PYG{p}{(}\PYG{n+nb+bp}{self}\PYG{p}{,} \PYG{n}{x}\PYG{p}{:} \PYG{n}{torch}\PYG{o}{.}\PYG{n}{Tensor}\PYG{p}{)} \PYG{o}{\PYGZhy{}}\PYG{o}{\PYGZgt{}} \PYG{n}{torch}\PYG{o}{.}\PYG{n}{Tensor}\PYG{p}{:}
        \PYG{c+c1}{\PYGZsh{} x.shape: [B, S, H, D]}
        \PYG{c+c1}{\PYGZsh{}}
        \PYG{c+c1}{\PYGZsh{} B: batch size}
        \PYG{c+c1}{\PYGZsh{} H: number of heads}
        \PYG{c+c1}{\PYGZsh{} S: sequence length}
        \PYG{c+c1}{\PYGZsh{} D: per head embedding size}
        \PYG{n}{x} \PYG{o}{=} \PYG{n+nb+bp}{self}\PYG{o}{.}\PYG{n}{phi}\PYG{p}{(}\PYG{n}{x}\PYG{p}{)}

        \PYG{c+c1}{\PYGZsh{} negative mapping enabled as in hedgehog}
        \PYG{n}{x} \PYG{o}{=} \PYG{n}{torch}\PYG{o}{.}\PYG{n}{cat}\PYG{p}{(}\PYG{p}{[}\PYG{n}{x}\PYG{p}{,} \PYG{o}{\PYGZhy{}}\PYG{n}{x}\PYG{p}{]}\PYG{p}{,} \PYG{n}{dim}\PYG{o}{=}\PYG{o}{\PYGZhy{}}\PYG{l+m+mi}{1}\PYG{p}{)}  \PYG{c+c1}{\PYGZsh{} [B, H, S, 2D]}

        \PYG{c+c1}{\PYGZsh{} NOTE: we use softmax as activation function here instead of}
        \PYG{c+c1}{\PYGZsh{} default exponential following hedgehog paper appendix to}
        \PYG{c+c1}{\PYGZsh{} avoid numerical overflows; softmax is applied on embedding}
        \PYG{c+c1}{\PYGZsh{} dimension here NOT sequence length as in standard softmax attention}
        \PYG{k}{return} \PYG{n}{x}\PYG{o}{.}\PYG{n}{softmax}\PYG{p}{(}\PYG{n}{dim}\PYG{o}{=}\PYG{o}{\PYGZhy{}}\PYG{l+m+mi}{1}\PYG{p}{)}
\end{Verbatim}
\caption{Implementation of the Hedgehog projection layer for Softmax Attention linearization (see \cref{sec:recipe:stage1} for details).}
\label{code::hedgehog}
\end{listing}

\end{document}